\documentclass{article}
\usepackage{placeins} 
\usepackage{microtype}
\usepackage{graphicx}
\usepackage{subcaption}
\usepackage{booktabs} 

\usepackage{multirow}
\usepackage{colortbl}
\usepackage{enumitem}
\usepackage{tikz}
\usepackage[table]{xcolor}

\definecolor{lightgray}{gray}{0.95}
\definecolor{darkgray}{gray}{0.85}

\newcommand{\bcirc}[1]{%
  \tikz[baseline=(char.base)]{
    \node[shape=circle, fill=black, inner sep=1.2pt] (char)
    {\textcolor{white}{\scriptsize #1}};
  }%
}
\newcommand{\down}[1]{\,{\scriptsize\textcolor{gray}{$\downarrow$#1}}}
\newcommand{\up}[1]{\,{\scriptsize\textcolor{gray}{$\uparrow$#1}}}
\usepackage{titletoc}

\usepackage[colorlinks, linkcolor=darkred, urlcolor=darkred, citecolor=darkred]{hyperref}
\usepackage{xcolor}
\definecolor{darkred}{RGB}{160, 0, 0} 

\usepackage{hyperref}

\usepackage[preprint]{icml2026}

\usepackage{amsmath}
\usepackage{amssymb}
\usepackage{mathtools}
\usepackage{amsthm}

\usepackage[capitalize,noabbrev]{cleveref}

\theoremstyle{plain}

\theoremstyle{definition}

\theoremstyle{remark}

\usepackage[textsize=tiny]{todonotes}

\begin{document}

\twocolumn[
  \icmltitle{Choosing How to Remember: Adaptive Memory Structures for LLM Agents}



  \icmlsetsymbol{equal}{*}

\begin{icmlauthorlist}
    \icmlauthor{Mingfei Lu}{uts}
    \icmlauthor{Mengjia Wu}{uts}
    \icmlauthor{Feng Liu}{unimelb}
    \icmlauthor{Jiawei Xu}{uta}
    \icmlauthor{Weikai Li}{ucla}
    \icmlauthor{Haoyang Wang}{uta}
    \icmlauthor{Zhengdong Hu}{uts}
    \icmlauthor{Ying Ding}{uta}
    \icmlauthor{Yizhou Sun}{ucla}
    \icmlauthor{Jie Lu}{uts}
    \icmlauthor{Yi Zhang}{uts}
\end{icmlauthorlist}

\icmlaffiliation{uts}{University of Technology Sydney, Sydney, Australia}
\icmlaffiliation{uta}{University of Texas at Austin, Austin, United States}
\icmlaffiliation{ucla}{University of California, Los Angeles, Los Angeles, United States}
\icmlaffiliation{unimelb}{The University of Melbourne, Melbourne, Australia}

\icmlcorrespondingauthor{Yi Zhang}{yi.zhang@uts.edu.au}

  \icmlkeywords{Machine Learning, ICML，Multiagent System, Memory Management}

  \vskip 0.3in
]



\printAffiliationsAndNotice{\icmlEqualContribution}

\begin{abstract}
Memory is critical for enabling large language model (LLM)–based agents to maintain coherent behavior over long-horizon interactions.
However, existing agent memory systems suffer from two key gaps:
they rely on a one-size-fits-all memory structure and do not model memory structure selection as a context-adaptive decision, limiting their ability to handle heterogeneous interaction patterns and resulting in suboptimal performance.
We propose a unified framework, \textsc{FluxMem}, that enables adaptive memory organization for LLM agents.
Our framework equips agents with multiple complementary memory structures.
It explicitly learns to select among these structures based on interaction-level features, using offline supervision derived from downstream response quality and memory utilization.
To support robust long-horizon memory evolution, we further introduce a three-level memory hierarchy and a Beta Mixture Model–based probabilistic gate for distribution-aware memory fusion, replacing brittle similarity thresholds.
Experiments on two long-horizon benchmarks, PERSONAMEM and LoCoMo, demonstrate that our method achieves average improvements of 9.18\% and 6.14\%.

\end{abstract}

\section{Introduction}

Large language model (LLM)–based agents have shown strong performance in interactive settings, including multi-step planning and open-ended dialogue~\citep{schlegel2025large,wang2024survey,li2024personal}. li2025survey,jia-etal-2025-hetgcot
In long-horizon interactions, agents must accumulate, retain, and retrieve information across many turns while user goals and contexts evolve over time~\citep{maharana2024evaluating}.
In this process, agent memory is responsible for recalling relevant information from rich, multi-turn user interactions. It directly influences the coherence of agents’ responses and long-term behavioral consistency. Therefore, effective memory organization is essential~\citep{zhang2025personaagentlargelanguagemodel,pan2025secom}.

Prior work on long-horizon memory for LLM agents falls into three lines.
(1) \textit{Flat retrieval-based memory systems} store salient facts or summaries in weakly structured stores and retrieve them via similarity search, offering efficiency but limited abstraction and relational reasoning~\citep{chhikara2025mem0,fang2025lightmem,zhong2024memorybank,gutierrezrag}.
(2) \textit{Explicitly structured memory systems} organize memory into linked notes or graph-like representations, enabling richer relational and temporal reasoning but typically relying on fixed organizational schemes~\citep{xu2025single,xu2025mem,wang2025mem,wang2025mirix,anonymous2026remem,rasmussen2025zep}.
(3) \textit{Policy managed memory systems} regulate memory storage and retrieval through higher-level control mechanisms.
These mechanisms are either learned (e.g., reinforcement learning–based controllers) or engineered (e.g., OS-inspired designs such as decay and redundancy filtering).
~\citep{yan2025memory,agrawal2025memory,kang2025memory,jonelagadda2025mnemosyne,langchain2025langmem,du2025memr,li2025memos}.

\begin{figure}[t]
  \centering
  \includegraphics[width=\linewidth]{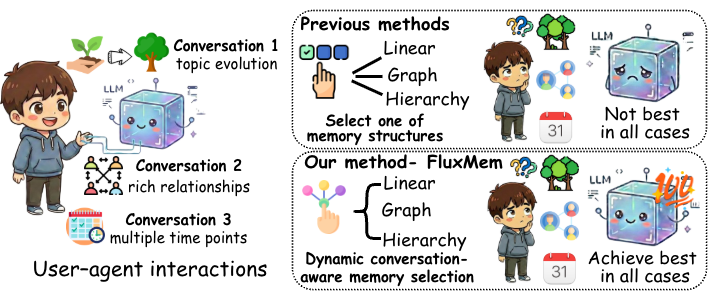}
  \caption{User–agent interactions exhibit diverse structural patterns, including relational, temporal, and topical variations.
Prior methods rely on a single memory structure and struggle across interaction types.
Our approach dynamically selects memory structures based on the conversation, enabling robust performance.
}
  \label{fig1}
\vskip -0.2in
\end{figure}

As illustrated in Fig.~\ref{fig1},  existing methods often struggle to deliver accurate and consistent responses in complex, long-horizon interaction scenarios, due to the following two research gaps.
\textbf{Gap 1: Single-structure assumption within memory systems.} 
Most existing methods~\citep{kang2025memory,rasmussen2025zep} assume that a single memory structure (e.g., graph) is sufficient across various tasks. 
Here, a memory structure refers to how the content within an individual memory unit (i.e., a compact dialogue segment) is organized and accessed, whereas relationships between different memory units concern how separate memories are linked or associated with one another.
This one-size-fits-all memory selection leads to suboptimal performance, as different tasks exhibit distinct structural characteristics and interaction patterns~\citep{trach2025mental,cellier2022dynamics,zhang2025survey}, such as topic evolution, temporal progression, and relational dependencies, that often co-occur in long-horizon conversations.
\textbf{Gap 2: Lack of conversation-adaptive memory structure selection.} 
While Gap 1 concerns how memory structure is designed, Gap 2 focuses on how memory structure is used during interaction.
Although some methods employ different memory structures across memory layers (e.g., linear for short-term and graph for long-term)~\citep{wang2025mem}, each layer remains fixed and non-adaptive to interaction content.
As a result, structure choice is excluded from optimization, limiting alignment with downstream performance and robustness under distribution shifts.

To bridge these two gaps, we propose a novel framework, \textsc{FluxMem}, that dynamically selects memory structures for LLM agents.
We identify a previously underexplored blind spot in existing memory systems:
\textbf{Memory structure is treated as a fixed design choice, rather than being adaptively adjusted based on interaction content.}

\textbf{To tackle Gap~1}, we move beyond the assumption of a single universal memory structure and equip agents with multiple complementary memory structures, each inducing distinct retrieval behaviors over the same interaction history.
This expands the expressiveness of agent memory and enables structural flexibility across heterogeneous tasks.
\textbf{To overcome Gap~2}, we elevate memory structure selection to an explicit, learnable mechanism.
We optimize the structure selection decision using offline, interaction-derived feedback obtained from the Multi-Session Chat (MSC) dataset~\citep{xu2022beyond}, aligning memory structures with downstream response quality and retrieval effectiveness.
In addition, we address a common practice in memory systems: using fixed similarity thresholds to decide whether a newly retrieved memory should be fused with an existing one.
These thresholds are usually set by hand and assume that similarity scores follow a stable pattern, which makes them sensitive to noise and unreliable when interaction distributions change.
Instead, we treat memory integration as a probabilistic gating process and propose a new Beta Mixture Model (BMM)–based gate that models the distribution of similarity scores directly.
By adaptively separating useful memories from noise, our approach removes the need for fixed thresholds and improves robustness in long-horizon interactions.
Extensive experiments on LoCoMo and PERSONAMEM demonstrate that our approach consistently outperforms existing memory systems in response accuracy and cross-turn consistency across long-horizon interaction tasks.
Our contributions are summarized as follows:

\bcirc{1}  \textbf{Key Observations.}
    We identify two gaps: reliance on a single memory structure and the absence of feedback-driven structure selection.

\bcirc{2} \textbf{Adaptive Multi-Structure Memory Framework.} 
    We propose a unified framework that supports multiple complementary memory structures and integrates structure selection via interaction-derived feedback.

\bcirc{3} \textbf{Memory Fusion and Update Mechanisms.} 
    We introduce mechanisms for memory fusion and update, including a BMM-based probabilistic gate.

\section{Related Work}
\label{sec:related_work}

LLMs have been widely applied across diverse domains, including dialogue systems, decision-making agents, code generation, and scientific reasoning\citep{yangadapting,li2025survey,jia-etal-2025-hetgcot,CHEN2026104336,wu2025leveraging,lu2025newbornimpactbiasawarecitation,jiang2025towards}. Their strong in-context learning and generative capabilities make them a natural foundation for building autonomous agent systems that require long-horizon interaction and memory management. Recent research on long-term memory for LLM agents can be grouped into three paradigms: (1) flat retrieval-based systems, (2) explicitly structured systems, and (3) policy-managed memory systems.

\textbf{Flat retrieval-based memory systems} store salient facts, events, or summaries in flat/weakly structured repositories and rely on similarity search for recall. Mem0, LightMem, and MemoryBank emphasize efficient extraction, incremental updating, and personalization~\citep{chhikara2025mem0,fang2025lightmem,zhong2024memorybank}. Their main limitation is that flat memory offers limited abstraction and relational reasoning, which can hurt performance on tasks requiring multi-hop inference or cross-context integration.

\textbf{Explicitly structured memory systems} organize memory into linked notes or graphs to support richer reasoning. A-MEM and O-Mem maintain evolving networks of notes via reflection and linking~\citep{xu2025mem,wang2025mem}; MemGAS further introduces multi-granularity memory with GMM-based organization and entropy routing~\citep{xu2025single}. MIRIX partitions memory into specialized types managed by multiple agents~\citep{wang2025mirix}, while Zep builds a temporally aware knowledge graph for cross-session consistency~\citep{rasmussen2025zep}. While effective for structure-aware recall, these methods rely on fixed memory structures or routing strategies, limiting adaptability across conversational patterns and reasoning demands.

\textbf{Policy managed memory systems} optimize memory behavior via learned policies or OS-inspired mechanisms. RL-based Memory-R1 and MIRA learn what to write/keep based on long-horizon rewards~\citep{yan2025memory,agrawal2025memory}. MemoryOS adopts concepts such as hierarchical storage and eviction~\citep{kang2025memory}, and Mnemosyne targets edge settings with redundancy pruning and probabilistic recall~\citep{jonelagadda2025mnemosyne}. Despite improved adaptivity, most approaches still assume fixed structural biases or hand-crafted organization rules, motivating methods that treat memory structure as a decision variable.

\begin{figure*}[h]
  \centering
  \includegraphics[width=\linewidth]{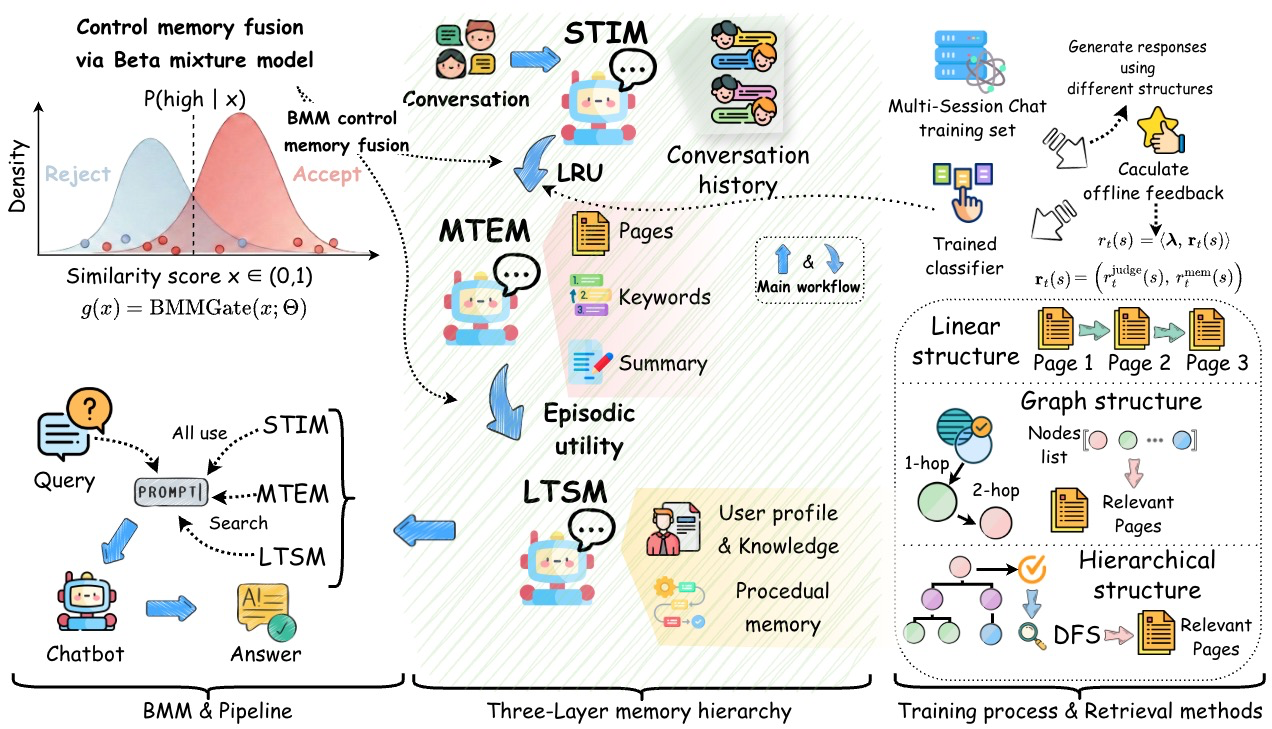}
  \caption{Overall architecture of \textit{FluxMem}.
The framework consists of three memory layers: STIM for buffering recent context, MTEM for structure-aware episodic storage, and LTSM for long-term semantic consolidation.
The central part depicts the \textbf{main workflow}, including memory writing, structure selection, and query-time retrieval across memory layers.}
  \label{fig2}
\end{figure*}

\section{Methodology}
\label{sec:methodology}

In this section, we introduce our framework, \textsc{FluxMem}, as illustrated in Fig.~\ref{fig2}. 
It comprises a three-layer memory hierarchy to support long-horizon interactions and a BMM-based gating mechanism controls memory fusion and updates. In mid-term memory, multiple complementary structures are supported, with a conversation-aware selector that dynamically chooses the most suitable structure based on context.
In the following parts, we first formally define the problem setting (\S3.1), then describe the three-layer memory hierarchy (\S3.2), multi-structure memory organization and retrieval (\S3.3), conversation-aware memory structure selection (\S3.4), and BMM-based memory fusion (\S3.5).

\subsection{Formalizaion}

We consider a long-horizon conversational setting where an agent interacts with a user over multiple turns.
Let $H_t = \{p_1, \dots, p_t\}$ denote the dialogue history up to turn $t$, where each \emph{page}
$p_i = (u_i, a_i)$ corresponds to a single user--agent exchange.
Given the current user query $q_t$, the goal is to generate a response $y_t$ by leveraging both the recent context and accumulated long-term conversational memories.

\paragraph{Memory Granularity and Hierarchy.}
We organize memory using a three-layer hierarchy operating at different temporal and semantic scales.
\emph{Short-term interaction memory} (STIM), denoted as $\mathcal{M}^S_t$, buffers the most recent pages to support fast recency-aware access.
\emph{Mid-term episodic memory} (MTEM), denoted as $\mathcal{M}^E_t = \{e_1, \dots, e_N\}$, stores a set of episodic memory units, where each unit
\begin{equation}
  e_j = \{p_{j,1}, \dots, p_{j,m}\},
\end{equation}
corresponds to an episodic session composed of multiple semantically or temporally related pages.
\emph{Long-term semantic memory} (LTSM), denoted as $\mathcal{M}^L$, maintains consolidated and durable knowledge abstracted from past interactions.

\paragraph{Memory Write and Consolidation Pipeline.}
Newly generated pages are first written into short-term memory and then progressively organized and consolidated across memory layers.
Formally, the memory update process can be abstracted as a stage-wise mapping:
\begin{equation}
p_i \xrightarrow{\phi_S} \mathcal{M}^S_t
\;\;\rightarrow\;\;
\mathcal{M}^E_t
\;\;\xrightarrow{\phi_L}\;\;
\mathcal{M}^L,
\end{equation}
where $\phi_S$ and $\phi_L$ denote abstract write and consolidation operators.
This hierarchical pipeline enables the system to balance fast access to recent interactions with structured storage of episodic history and long-term semantic retention.

\paragraph{Structured Episodic Memory.}
Each episodic memory unit in MTEM is assigned a single suitable structure (linear, graph, or hierarchical) based on its conversational content, which determines how the unit is organized and retrieved.
The selection of structure is treated as a context-dependent decision, allowing different episodic units to be organized under different structural forms.

\paragraph{Query and Response Generation.}
Given a query $q_t$, the agent retrieves and fuses relevant information across memory layers to form a memory-conditioned context:
\begin{equation}
\tilde{\mathcal{M}}_t = \Psi(q_t, \mathcal{M}^S_t, \mathcal{M}^E_t, \mathcal{M}^L),
\end{equation}
where $\Psi(\cdot)$ denotes a memory retrieval and fusion function.
The final response is generated by conditioning on both the query and the retrieved memories:
\begin{equation}
y_t \sim p(y \mid q_t, \tilde{\mathcal{M}}_t).
\end{equation}

\subsection{Three-Layer Memory Hierarchy}
To support long-horizon interactions, we organize memory into a three-level hierarchy: STIM, MTEM, and LTSM.

\paragraph{Short-Term Interaction Memory.}
STIM buffers the most recent dialogue history to provide fast access to highly recent context, which is usually most relevant to the user’s next query.
During retrieval, all contents in STIM are directly incorporated into the generation context without filtering. However, an unbounded short-term buffer introduces noise and destabilizes downstream memory organization, necessitating strict capacity control.

The design of STIM is motivated by findings from cognitive psychology~\citep{cowan2001magical, cowan2010magical, vogel2004neural, mathy2012s}, which show that human working memory can maintain only a few unstructured items, typically around three to five chunks.
Accordingly, we set the capacity of STIM to four pages to enforce a strict short-term memory budget.
When this limit is exceeded, older pages are irreversibly transferred to MTEM for structured storage using a recency-based Least Recently Used (LRU) policy, reflecting the dominant role of recency in short-term conversational relevance.

Formally, let $\mathcal{M}_t$ denote the set of interaction pages stored in STIM at turn $t$, and let $C$ be the fixed STIM capacity.
When $|\mathcal{M}_t| > C$, we evict a subset of pages
\begin{equation}
\mathcal{E}_t = \arg\min_{\mathcal{P} \subset \mathcal{M}_t,\; |\mathcal{P}| = |\mathcal{M}_t| - C}
\sum_{p \in \mathcal{P}} \tau(p),
\end{equation}
where $\tau(p)$ denotes the last access time of page $p$.
The selected pages $\mathcal{E}_t$ are then flushed to MTEM.

Importantly, transferring information to MTEM introduces a second, qualitatively different challenge.
Unlike STIM, MTEM organizes interaction history into episodic units, where indiscriminately appending every incoming page would lead to fragmented or redundant structures.
We therefore apply a probabilistic integration decision to determine whether a transferred page should be merged into an existing session and, if so, where it best fits.
This decision is implemented via a BMM, which guides integration using the distribution of similarity scores rather than fixed thresholds.
Details are provided in Section~\S3.5.

\paragraph{Mid-Term Episodic Memory.}
MTEM serves as the primary repository for structured interaction history and bridges transient short-term context with long-term semantic memory.
A key challenge at this stage is controlling the granularity and growth of stored information: storing individual interaction pages in isolation leads to fragmented memory representations and noisy retrieval, while indiscriminate accumulation results in unbounded memory expansion.

To address this challenge, MTEM organizes interaction history into episodic units, each grouping semantically related interaction pages into a coherent unit that serves as the basic unit for memory organization and retrieval.
Each session contains (i) a sequence of pages with dialogue content, timestamps, representations, and continuity links; (ii) session-level metadata, including a summary and its representations; and (iii) indexing structures (e.g., entity--relation graphs or topic hierarchies) corresponding to the assigned memory structure.
As MTEM grows, only a subset of units remains consistently useful.
We therefore estimate the long-term utility of each session using a lightweight episodic utility score,
$U(s)=w_1 c(s) + w_2 \ell(s) + w_3 d(s)$,
which combines access frequency, interaction intensity, and recency.
Higher-utility units are prioritized for consolidation into LTSM, keeping MTEM compact and focused.

\paragraph{Long-Term Semantic Memory.}
While MTEM preserves rich episodic interaction history, repeatedly retrieving episodic units is inefficient for information that remains stable across interactions.
LTSM therefore consolidates high-utility episodic experiences into durable semantic knowledge for long-term reuse.
Unlike episodic memory, LTSM stores abstracted and persistent information that is largely invariant to interaction context, such as user profiles, user-specific facts, general knowledge, and reusable interaction strategies.
To maintain long-term memory under capacity constraints, LTSM employs an eligibility-based pruning mechanism rather than heuristic scalar scoring:
\begin{equation}
m \in \mathcal{M}_{\text{LTSM}} \;\; \text{iff} \;\;
u(m) \ge \tau_u \;\land\; r(m) \ge \tau_r \;(\land\; c(m) \ge \tau_c),
\end{equation}
where $u(m)$, $r(m)$, and $c(m)$ denote usage, recency, and optional confidence, respectively, and $\tau_*$ are the corresponding thresholds. 
Items failing to meet these criteria are removed when capacity is exceeded, ensuring compact and reliable long-term storage.
Together, STIM, MTEM, and LTSM form a hierarchical memory stack that separates transient context, episodic history, and stabilized semantic knowledge, enabling dynamic memory organization.

\subsection{Multi-Structure Organization and Retrieval}
\label{sec:multi_structure}

Episodic units in MTEM are heterogeneous and best organized under different memory structures.
Some units benefit from temporal organization, while others are better captured by relational or hierarchical structures.
Therefore, imposing a single memory organization leads to retrieval errors across interaction scenarios.
To address this, our framework maintains multiple complementary memory structures~\citep{lewis2020retrieval,collins1969retrieval}.
\emph{From a structural perspective}, we restrict our design space to linear, graph-based, and hierarchical memory structures as they constitute the foundational primitives of information organization.
Classical data structure theory and information science show that complex information systems are fundamentally organized through sequential, hierarchical, or networked forms, while other structures can be reduced to or composed from these basic organizations~\citep{cormen2022introduction,hopcroft1983data,simon2012architecture,buckland1991information}.

\paragraph{Linear Memory.}
Linear memory organizes episodic content chronologically, preserving the order of information acquisition.
Retrieval is primarily driven by semantic similarity with an implicit recency effect, making this structure effective for temporally dependent queries such as step-by-step instructions or evolving goals.

\paragraph{Graph Memory.}
Graph memory organizes episodic content around relational structure.
Retrieval is entity-centric, combining neighborhood expansion with semantic matching, and favors queries where relevance is defined by relational proximity rather than time.

\paragraph{Hierarchical Memory.}
Hierarchical memory organizes episodic content into multi-level abstractions.
Retrieval follows a coarse-to-fine strategy by matching higher-level topics and then accessing finer-grained content, supporting abstraction-aware recall across granularity levels.

Each memory structure emphasizes a different aspect of conversational information, including temporal, relational, and hierarchical aspects, motivating explicit structure selection.
The next section describes how the agent dynamically selects among these structures based on the context.

\subsection{Context-Aware Memory Structure Selection}
\label{sec:structure_selection}

We formulate memory structure selection as a context-aware classification problem.
At each interaction turn $t$, the agent selects the memory structure that best matches the current conversational context and retrieval needs.
Let $\mathcal{S}=\{\textsc{Linear}, \textsc{Graph}, \textsc{Hierarchical}\}$ denote the set of supported structures.
Given the context, a lightweight selector predicts the most suitable structure from $\mathcal{S}$.

\paragraph{Conversation feature representation.}
We represent the conversation state using a compact set of interpretable interaction-level features, rather than dense embeddings or heuristic rules.
At each turn $t$, the interaction context is encoded as a feature vector $x_t \in \mathbb{R}^{d}$ capturing structural cues relevant to retrieval, including interaction scale, temporal signals, entity--relation patterns, and topic structure.
This lightweight representation enables efficient structure selection without additional reasoning overhead.

\paragraph{Structure selector.}
The selector is implemented as a shallow multi-layer perceptron (MLP) classifier that outputs a score for each candidate structure,
\begin{equation}
f_\theta(x_t) = \mathrm{Softmax}\big(g_\theta(x_t)\big),
\end{equation}
where $g_\theta(\cdot)$ denotes the MLP.
The predicted structure is obtained as
\begin{equation}
\hat{s}_t = \arg\max_{s \in \mathcal{S}} f_\theta(x_t)[s].
\end{equation}
The selected structure $\hat{s}_t$ is then used to organize episodic units in MTEM and guide subsequent retrieval.

\paragraph{Offline supervision via interaction-derived rewards.}
Since ground-truth labels for optimal memory structures are not directly available, we construct supervision signals offline using interaction-derived rewards.
For each training interaction, all candidate structures are evaluated by running the agent under structure $s$ and computing a scalar reward
\begin{equation}
r_t(s) = \lambda_q \, r_t^{\text{judge}}(s) + \lambda_m \, r_t^{\text{mem}}(s),
\end{equation}
where $r_t^{\text{judge}}$ denotes a response quality score and $r_t^{\text{mem}}$ measures memory utilization effectiveness.
The structure achieving the highest reward, $s_t^{\ast} = \arg\max_{s \in \mathcal{S}} r_t(s)$, is treated as the target label for that interaction.

The selector is trained offline by minimizing a standard supervised classification loss,
\begin{equation}
\mathcal{L}_{\text{sel}}(\theta)
= \sum_t \ell\big(f_\theta(x_t),\, s_t^{*}\big),
\end{equation}
where $\ell(\cdot,\cdot)$ denotes the cross-entropy loss.

At inference stage, the trained selector predicts the most suitable memory structure for the current interaction context.
As the conversation evolves, structure selection is performed repeatedly, allowing memory organization to adapt dynamically to changing interaction patterns.

\subsection{Beta-Mixture-Gated Memory Fusion}
\label{sec:bmm_gate}

In long-horizon interactions, deciding whether newly arriving information should be merged into existing memory is unstable.
Fixed similarity thresholds or heuristic top-$K$ rules are brittle, as similarity score distributions vary across conversation types and interaction stages.
To address this issue, we adopt a \emph{BMM–based gating mechanism} that provides a soft, distribution-aware criterion for memory fusion.

Given an incoming interaction summary and a set of candidate episodic units, we compute a matching score for each candidate and normalize it to the unit interval $x \in (0,1)$.
These normalized scores are modeled using a two-component Beta mixture:
\begin{equation}
p(x)=\pi\,\mathrm{Beta}(x;\alpha_1,\beta_1)
      +(1-\pi)\,\mathrm{Beta}(x;\alpha_0,\beta_0),
\end{equation}
where two components correspond to high- and low-compatibility score regimes.
Here, $\pi$ denotes the mixture weight, and $(\alpha_k,\beta_k)$ parameterize the Beta distribution of component $k$.
Mixture parameters are estimated via an EM procedure with responsibility-weighted moment updates.
Based on the mixture, we compute the posterior probability,
\begin{equation}
g(x)=P(z=1\mid x)
=\frac{\pi\,\mathrm{Beta}(x;\alpha_1,\beta_1)}
{\sum_{k\in\{0,1\}}\pi_k\,\mathrm{Beta}(x;\alpha_k,\beta_k)},
\end{equation}
where $z\in\{0,1\}$ is a latent component indicator and $z=1$ denotes the high-compatibility component.
The posterior $g(x)$ serves as a soft gating signal that measures how likely a candidate belongs to the high-compatibility regime under the learned score distribution.

Importantly, fusion decisions are made in the posterior probability space rather than on raw similarity scores.
Candidates with high posterior mass under the high-compatibility component are retained, with a minimum-keep safeguard to avoid over-filtering.
Retained candidates are merged into the most compatible episodic unit under the selected memory structure, or stored as a new unit if none qualify.
By operating at the distribution level, the BMM gate replaces brittle hand-tuned thresholds with an adaptive fusion criterion.

\section{Experiments}
\label{sec:experiments}

\subsection{Experimental Setup}
\subsubsection{Datasets}
To evaluate the effectiveness of our proposed framework, we conduct a series of experiments across two representative datasets - LoCoMo~\citep{maharana2024evaluating} and PERSONAMEM~\citep{jiang2025know}. Both of these datasets are designed to test long-horizon reasoning and memory capabilities in LLM-based agents.

\subsubsection{Baselines}
We compare \textit{FluxMem} against a range of state-of-the-art memory systems, including: \textbf{LangMem}~\citep{langchain2025langmem}: A language model with built-in memory capabilities.
\textbf{Mem0}~\citep{chhikara2025mem0}: A flat retrieval-based memory system.
\textbf{ZEP}~\citep{rasmussen2025zep}: A time-aware graph memory system.
\textbf{MemoryOS}~\citep{kang2025memory}: An OS-inspired memory management system.
\textbf{A-Mem}~\citep{xu2025mem}: An explicit structured memory system using linked notes.
\textbf{O-mem}~\citep{wang2025mem}: A graph-based memory organization method.
\textbf{MemR$^3$}~\citep{du2025memr}: A closed-loop controller that dynamically alternates between retrieval, reflection, and answering to optimize evidence coverage for LLM agents.
\textbf{MEMOS}~\citep{li2025memos}: A redundancy-filtering memory system for edge settings.
\textbf{HippoRAG 2}~\citep{gutierrezrag}: A framework that uses knowledge graphs for efficient retrieval.

\begin{table*}[t]
\centering
\caption{Category-wise performance on PERSONAMEM. Best results in each column are shown in bold, second-best are underlined.}
\small
\setlength{\tabcolsep}{3pt}
\begin{tabular}{l|c|c|c|c|c|c|c}
\toprule
Method &
\begin{tabular}[c]{@{}c@{}}Recall user\\ shared facts\end{tabular} &
\begin{tabular}[c]{@{}c@{}}Suggest new\\ ideas\end{tabular} &
\begin{tabular}[c]{@{}c@{}}Track full\\ preference\\ evolution\end{tabular} &
\begin{tabular}[c]{@{}c@{}}Revisit reasons\\ behind\\ preference updates\end{tabular} &
\begin{tabular}[c]{@{}c@{}}Provide preference-\\aligned\\ recommendations\end{tabular} &
\begin{tabular}[c]{@{}c@{}}Generalize to\\ new scenarios\end{tabular} &
\textbf{Average} \\
\midrule
LangMem   & 31.29 & 24.73 & 53.24 & 81.82 & 40.00 &  8.77 & 40.64 \\
Mem0     & 32.13 & 15.05 & 54.68 & 80.81 & 52.73 & 57.89 & 48.55 \\
A-Mem    & 63.01 & 27.96 & 54.68 & 85.86 & 69.09 & 57.89 & 59.75 \\
Memory OS & 72.72 & 17.20 & 58.27 & 78.79 & \textbf{72.72} & 56.14 & 59.97 \\
O-Mem & 67.81 & 21.51 & \underline{61.15} & \underline{89.90} & 65.45 & \underline{73.68} & 63.25 \\
HippoRAG 2 & \underline{76.74} & \underline{33.33} & 56.12 & 78.79 & 23.64 & 38.60 & 51.87 \\
\rowcolor{lightgray}
\textbf{FluxMem} & \textbf{85.27} & \textbf{36.56} & \textbf{65.47} & \textbf{91.92} & \underline{70.91} & \textbf{82.46} & \textbf{72.43} \\
\midrule
\rowcolor{darkgray}
\textbf{Improvement (\%)}
& +8.53 & +3.23 & +4.32 & +2.02 & -1.81 & +8.78 & +9.18 \\
\bottomrule
\end{tabular}
\label{tab:personamem_llm}
\end{table*}

\begin{table*}[t]
  \centering
  \caption{Performance comparison on LoCoMo. Best results in each column are shown in bold, second-best are underlined.}
  \small
  \setlength{\tabcolsep}{2.5pt}
  \begin{tabular}{l|cccc|cccc|cccc|cccc|ccc}
    \toprule
    \multirow{2}{*}{Method} 
    & \multicolumn{4}{c|}{Cat1: Multi-hop} 
    & \multicolumn{4}{c|}{Cat2: Temporal} 
    & \multicolumn{4}{c|}{Cat3: Open} 
    & \multicolumn{4}{c|}{Cat4: Single-hop}
    & \multicolumn{3}{c}{\textbf{Average}} \\
    & F1$\uparrow$ & B-1$\uparrow$ & R-L$\uparrow$ &  
    & F1$\uparrow$ & B-1$\uparrow$ & R-L$\uparrow$ &  
    & F1$\uparrow$ & B-1$\uparrow$ & R-L$\uparrow$ &  
    & F1$\uparrow$ & B-1$\uparrow$ & R-L$\uparrow$ &  
    & F1$\uparrow$ & B-1$\uparrow$ & R-L$\uparrow$ \\
    \midrule
    LangMem  
    & 41.11 & 32.09 & 37.71 & 
    & 53.67 & 46.22 & 47.37 & 
    & \underline{33.38} & 27.26 & 24.55 & 
    & 51.13 & 44.22 & 48.37 & 
    & 44.32 & 37.95 & 39.50 \\

    Mem0     
    & 28.84 & 20.22 & 27.45 & 
    & 24.21  & 18.75 & 23.23 & 
    & 23.59 & 19.09 & 23.13 & 
    & 32.74 & 27.58 & 33.02 & 
    & 27.35 & 21.41 & 26.21 \\

    ZEP      
    & 8.74  & 5.98 & 6.79 & 
    & 7.40  & 4.70 & 6.07 & 
    & 7.64  & 3.46 & 5.18 & 
    & 7.91  & 3.80 & 6.26 & 
    & 7.92 & 4.48 & 6.08 \\

    MemoryOS
    & \underline{43.03} & \underline{34.09} & \underline{39.75} & 
    & 44.43 & 32.03 & 42.64 & 
    & 28.36 & 23.24 & \underline{27.83} & 
    & 49.31 & 42.61 & \underline{49.84} & 
    & 41.28 & 32.99 & \underline{40.02} \\

    A-Mem   
    & 32.41 & 24.28 & 31.26 & 
    & 37.23 & 33.02 & 39.61 & 
    & 16.16 & 14.64 & 15.19 & 
    & 40.94 & 35.54 & 42.32 & 
    & 31.69 & 26.87 & 32.10 \\

    MEMOS   
    & 20.27 & 13.35 & 15.75 & 
    & 14.86 & 7.86 & 11.58 & 
    & 11.87 & 6.85 & 9.32 & 
    & 18.96 & 11.42 & 17.37 & 
    & 16.49 & 9.87 & 13.51 \\

    MemR$^{3}$   
    & 18.65 & 11.17 & 14.37 & 
    & 25.39 & 15.90 & 22.63 & 
    & 11.56 & 5.77 & 8.96 & 
    & 24.35 & 12.72 & 22.00 & 
    & 20.49 & 11.89 & 17.49 \\

    O-mem  
    & 42.64 & 34.08 & 32.78 & 
    & \textbf{57.48} & \textbf{49.76} & \underline{48.88} & 
    & 30.58 & \underline{25.69} & 21.11 & 
    & \underline{54.89} & \underline{48.98} & 44.26 & 
    & \underline{46.40} & \underline{39.63} & 36.76 \\

    HippoRAG 2
    & 28.11 & 18.92 & 27.68 & 
    & 10.35 & 6.92 & 9.82 & 
    & 24.91 & 21.10 & 24.48 & 
    & 44.82 & 37.25 & 44.37 & 
    & 27.55 & 21.05 & 26.59 \\
\rowcolor{lightgray}
    \textbf{FluxMem} 
    & \textbf{48.56} & \textbf{39.90} & \textbf{44.05} & 
    & \underline{56.67} & \underline{41.88} & \textbf{54.89} & 
    & \textbf{37.30} & \textbf{29.63} & \textbf{36.76} & 
    & \textbf{62.12} & \textbf{53.52} & \textbf{62.33} & 
    & \textbf{51.16} & \textbf{41.73} & \textbf{49.51} \\
    \midrule
    \rowcolor{darkgray}
    \textbf{Improvement (\%)} 
      & +5.53 & +5.81 & +4.3 & 
      & -0.81 & -7.68 & +6.01 & 
      & +3.92 & +3.94 & +8.93 & 
      & +7.23 & +4.54 & +12.49 & 
      & +6.84 & +2.1 & +9.49 \\
    \bottomrule
  \end{tabular}
  \label{tab:locomo_main}
\end{table*}

\begin{table*}[t]
\centering
\caption{Ablation study on PERSONAMEM.}
\small
\setlength{\tabcolsep}{4pt}
\begin{tabular}{lcccccc}
\toprule
Method &
\begin{tabular}[c]{@{}c@{}}Recall user\\ shared facts\end{tabular} &
\begin{tabular}[c]{@{}c@{}}Suggest new\\ ideas\end{tabular} &
\begin{tabular}[c]{@{}c@{}}Track full\\ preference\\ evolution\end{tabular} &
\begin{tabular}[c]{@{}c@{}}Revisit reasons\\ behind\\ preference updates\end{tabular} &
\begin{tabular}[c]{@{}c@{}}Provide preference-\\aligned\\ recommendations\end{tabular} &
\begin{tabular}[c]{@{}c@{}}Generalize to\\ new scenarios\end{tabular} \\
\midrule
\textit{w/o Linear}   
& 80.62\down{4.7} 
& \textbf{40.86}\up{4.3} 
& 62.59\down{2.9} 
& \underline{90.91}\down{1.0} 
& \underline{69.09}\down{1.8} 
& 75.44\down{7.0} \\

\textit{w/o Graph}      
& 78.29\down{7.0} 
& 34.41\down{2.2} 
& \underline{63.31}\down{2.2} 
& 86.87\down{5.1} 
& 67.27\down{3.6} 
& 75.44\down{7.0} \\

\textit{w/o Hierarchy}     
& 79.84\down{5.4} 
& 35.48\down{1.1} 
& 61.87\down{3.6} 
& \underline{90.91}\down{1.0} 
& 63.64\down{7.3} 
& \underline{78.95}\down{3.5} \\

\textit{w/o BMM}     
& \underline{81.40}\down{3.9} 
& \underline{37.63}\up{1.1} 
& 61.87\down{3.6} 
& 87.88\down{4.0} 
& \underline{69.09}\down{1.8} 
& \underline{78.95}\down{3.5} \\

\rowcolor{lightgray}
\midrule
\textbf{FluxMem} 
& \textbf{85.27} 
& 36.56 
& \textbf{65.47} 
& \textbf{91.92} 
& \textbf{70.91} 
& \textbf{82.46} \\
\bottomrule
\end{tabular}
\label{tab:personamem_ablation}
\end{table*}

\begin{table*}[t]
  \centering
\caption{Ablation study on LoCoMo}
\small
\setlength{\tabcolsep}{1pt}
\begin{tabular}{l|ccc|ccc|ccc|ccc}
\toprule
\multirow{2}{*}{Method} 
& \multicolumn{3}{c|}{Cat1: Multi-hop} 
& \multicolumn{3}{c|}{Cat2: Temporal} 
& \multicolumn{3}{c|}{Cat3: Open} 
& \multicolumn{3}{c}{Cat4: Single-hop} \\
& F1 & B-1 & R-L & F1 & B-1 & R-L & F1 & B-1 & R-L & F1 & B-1 & R-L \\
\midrule
\textit{w/o Linear}     
& \underline{44.39}\down{4.2} & \underline{36.63}\down{3.3} & \underline{36.73}\down{7.3} 
& 49.86\down{6.8} & 36.76\down{5.1} & 49.86\down{5.0}
& \underline{29.60}\down{7.7} & \underline{6.67}\down{23} & \underline{27.41}\down{9.4}
& 58.75\down{3.4} & 47.26\down{6.3} & 58.49\down{3.8} \\

\textit{w/o Graph}    
& 41.95\down{6.6} & 32.23\down{7.7} & 32.31\down{12}
& 53.95\down{2.7} & 41.38\down{0.5} & 53.95\down{0.9}
& 17.86\down{19} & 4.24\down{25} & 17.76\down{19}
& \underline{61.39}\down{0.7} & \underline{52.65}\down{0.9} & \underline{61.63}\down{0.7} \\

\textit{w/o Hierarchy}    
& 42.36\down{6.2} & 32.00\down{7.9} & 35.38\down{8.7}
& \underline{55.50}\down{1.2} & \textbf{43.97}\up{2.1} & \textbf{55.50}\up{0.6}
& 25.28\down{12} & 4.89\down{25} & 25.28\down{12}
& 58.39\down{3.7} & 47.89\down{5.6} & 59.23\down{3.1} \\

\textit{w/o BMM}    
& 41.18\down{7.4} & 31.70\down{8.2} & 32.48\down{12}
& 54.33\down{2.3} & 41.75\down{0.1} & 54.33\down{0.6}
& 20.97\down{16} & 3.79\down{26} & 20.97\down{16}
& 59.74\down{2.4} & 47.74\down{5.8} & 59.05\down{3.3} \\

\midrule
\rowcolor{lightgray}
\textbf{FluxMem} 
& \textbf{48.56} & \textbf{39.90} & \textbf{44.05}
& \textbf{56.67} & \underline{41.88} & \underline{54.89}
& \textbf{37.30} & \textbf{29.63} & \textbf{36.76}
& \textbf{62.12} & \textbf{53.52} & \textbf{62.33} \\
\bottomrule
\end{tabular}
\label{tab:locomo_ablation}
\end{table*}

\subsubsection{implementation Details}
Experiments are conducted on 2×A100 GPUs using GPT-4.1 with temperature 0, to ensure reproducible outputs.
We employ a three-layer memory: STIM (capacity 4, LRU), MTEM (up to 2000 episodic sessions with utility-aware pruning), and LTSM (up to 100 consolidated entries retrieved by semantic similarity).
Memory fusion in MTEM is governed by a two-component Beta Mixture Model gate, where matching scores are filtered by the high-compatibility posterior (threshold 0.6, minimum keep 1).
The memory structure selector is a two-layer MLP (12-dim input, hidden size 4) with a 3-way output over \textsc{Linear}/\textsc{Graph}/\textsc{Hierarchical}, trained with interaction-derived supervision.
Retrieval fuses dense encoding (all-MiniLM-L6-v2) and BM25 via reciprocal rank fusion.

\subsubsection{Evaluation Metrics}
We evaluate memory systems using dataset-specific metrics. On the \textsc{PersonaMem} dataset, where each query is formulated as a single-choice question, we use accuracy as the evaluation metric.
On the \textsc{LoCoMo} dataset, which involves free-form response generation, we adopt a multi-dimensional evaluation suite, including F1, BLEU-1 (B1), ROUGE-L, ROUGE-1, ROUGE-2, and BERTScore, to comprehensively assess response quality.

\subsection{Overall Performance}
Tables~\ref{tab:personamem_llm} and~\ref{tab:locomo_main} report the overall results on \textsc{PersonaMem} and \textsc{LoCoMo}.
Overall, \textbf{FluxMem} achieves the strongest and most consistent performance across both datasets, demonstrating its effectiveness for long-horizon conversational memory.
On \textsc{PersonaMem} (Table~\ref{tab:personamem_llm}), FluxMem attains the highest average accuracy (72.43\%), outperforming all baselines by +9.18\%, with clear advantages on preference-centric tasks such as tracking preference evolution, revisiting update reasons, and generalizing to new scenarios.
On \textsc{LoCoMo} (Table~\ref{tab:locomo_main}), FluxMem again delivers the best overall results, achieving the highest average F1, BLEU-1, and ROUGE-L scores, with particularly strong gains in multi-hop, open, and single-hop reasoning.
\begin{figure}[h]
  \centering
  \includegraphics[width=\linewidth]{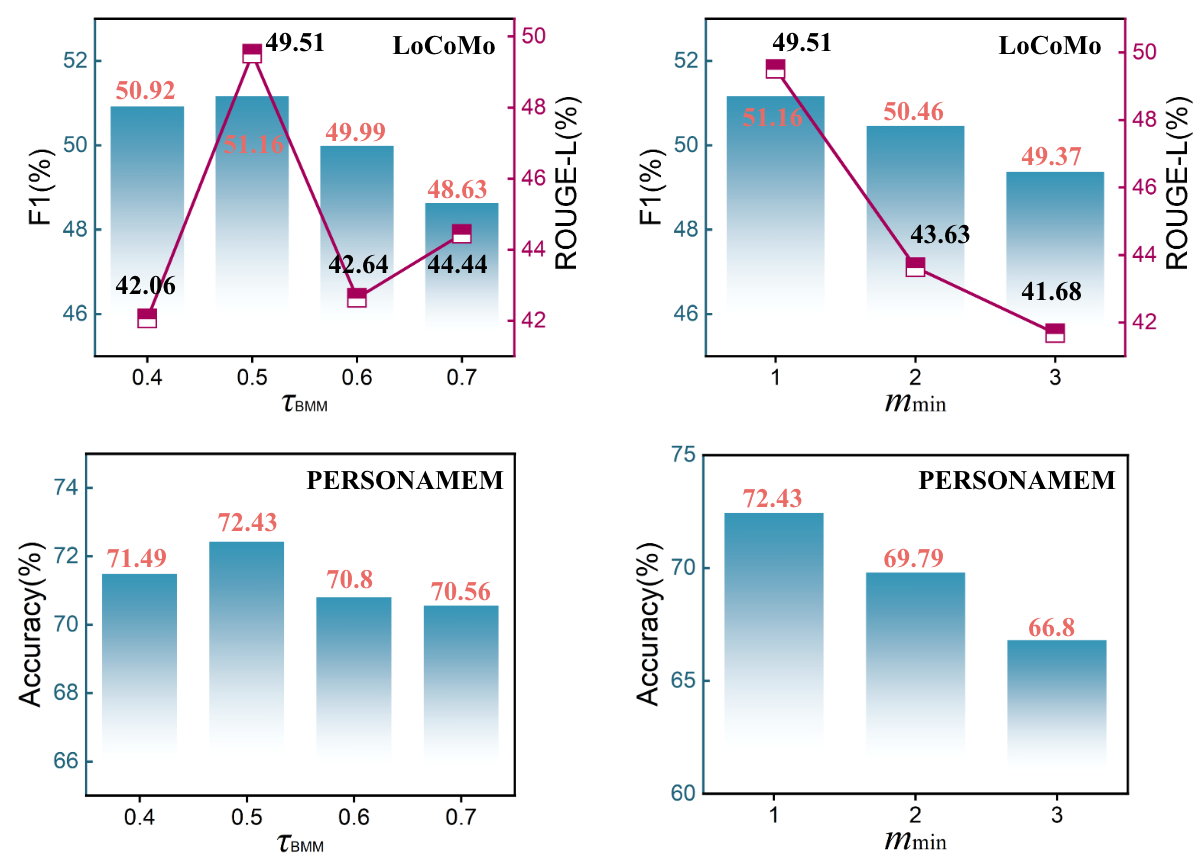}
  \caption{Parameter experiments results.}
  \label{fig:hyper}
\vskip -0.1in
\end{figure}
In contrast, competing methods relying on fixed memory structures or layer-specific heuristics perform well only in isolated settings and struggle to adapt to varying interaction patterns and retrieval requirements.
In a few categories, FluxMem does not outperform the strongest baselines, which is expected as certain tasks align closely with specific fixed structures.
This reflects an trade-off between specializing for individual scenarios and maintaining robust performance across diverse long-horizon interaction patterns.
Overall, these results show that multi-structure episodic memory with context-adaptive structure selection yields more stable and robust improvements than fixed-structure systems.

\subsection{Ablation Studies}
We conduct ablation studies on \textsc{PersonaMem} and \textsc{LoCoMo} to examine the contribution of individual components in FluxMem (Tables~\ref{tab:personamem_ablation} and~\ref{tab:locomo_ablation}).
Removing the linear structure causes the largest drop on temporally dependent tasks such as \emph{Track full preference evolution}, highlighting the importance of chronological organization.
Eliminating the graph or hierarchical structure mainly affects relational alignment and abstraction-oriented tasks, respectively.
Removing the BMM-based gating mechanism also leads to noticeable degradation across categories, indicating the importance of distribution-aware memory fusion.
On \textsc{LoCoMo}, all ablated variants underperform the full model, with clear drops in multi-hop and open reasoning.
Removing linear, graph, or hierarchical memory impairs temporally sensitive, relational, and abstraction-driven queries, respectively.
The pronounced degradation on the \textsc{Open} category is expected, as open-domain questions often combine weak temporal cues with sparse relational signals, making them particularly dependent on flexible structure selection and effective cross-structure memory fusion.
Notably, \textit{w/o BMM} consistently lags behind FluxMem, underscoring the benefit of adaptive gating over heuristic similarity thresholds.
Overall, these results show that FluxMem’s components are complementary: multi-structure memory captures diverse interaction patterns, while adaptive selection and BMM-based gating support robust long-horizon reasoning.

\subsection{Analysis of Parameter Sensitivity}
We analyze the sensitivity of two key hyper-parameters that directly control the selectivity of BMM-based memory fusion: the posterior threshold $\tau_{\text{BMM}}$ and the minimum retention parameter $m_{\min}$ (Fig.~\ref{fig:hyper}).
(1) $\tau_{\text{BMM}}$.
Fixing $m_{\min}=1$, we vary $\tau_{\text{BMM}}$ from 0.4 to 0.7.
On \textsc{PersonaMem}, accuracy peaks at $\tau_{\text{BMM}}=0.5$, while both lower and higher values degrade performance.
Smaller thresholds admit low-compatibility candidates, introducing noise, whereas larger thresholds become overly restrictive.
A similar trend is observed on \textsc{LoCoMo}, where overall F1, BLEU-1, and ROUGE-L peak at $\tau_{\text{BMM}}=0.5$ and decline as the threshold increases.
(2) $m_{\min}$.
Fixing $\tau_{\text{BMM}}=0.5$, we vary $m_{\min}$ from 1 to 3.
On both \textsc{PersonaMem} and \textsc{LoCoMo}, increasing $m_{\min}$ consistently degrades performance, indicating that forcing the retention of additional candidates introduces redundant or weakly relevant memories.
Overall, FluxMem shows moderate sensitivity, favoring $\tau_{\text{BMM}}=0.5$ and $m_{\min}=1$ for stable performance.

\subsection{Case Studies}
We present three representative cases (Figs.~\ref{figcase1}--\ref{figcase3}) to illustrate how FluxMem dynamically leverages linear, graph, and hierarchical memory structures for diverse long-horizon queries.
\textbf{Time-dependent reasoning.}
As shown in Fig.~\ref{figcase1}, the first case asks when Melanie read a book, requiring alignment of a relative temporal expression with a session timestamp.
FluxMem selects linear memory to preserve chronology and correctly infers the year (2022), while baselines fail due to missing temporal linkage.
\textbf{Complex relational reasoning.}
Fig.~\ref{figcase2} presents a query about Caroline’s place of origin, which requires integrating relocation history with a later country mention.
FluxMem activates graph memory to connect relational cues across sessions and correctly identifies Sweden, whereas competing methods fail to establish the necessary relations.
\textbf{Topic evolution.}
In Fig.~\ref{figcase3}, the query appears in a conversation with substantial topic drift.
By organizing episodic content hierarchically, FluxMem abstracts over evolving subtopics and retrieves the correct activity (horseback riding), avoiding interference from recent but irrelevant information.

\section{Conclusion}
We introduce FluxMem, a multi-structure conversational memory framework with adaptive structure selection for long-horizon interactions.
By integrating linear, graph, and hierarchical memories with conversation-aware selection and BMM-based fusion, FluxMem supports diverse reasoning demands.
Experiments on \textsc{PersonaMem} and \textsc{LoCoMo} demonstrate consistent improvements, highlighting the effectiveness of adaptive memory structuring.


\nocite{langley00}

\bibliography{main}

\begin{thebibliography}{45}
\providecommand{\natexlab}[1]{#1}
\providecommand{\url}[1]{\texttt{#1}}
\expandafter\ifx\csname urlstyle\endcsname\relax
  \providecommand{\doi}[1]{doi: #1}\else
  \providecommand{\doi}{doi: \begingroup \urlstyle{rm}\Url}\fi

\bibitem[Agrawal et~al.(2025)Agrawal, Kher, Mittal, Maheshwari, and Balasubramanian]{agrawal2025memory}
Agrawal, S., Kher, K.~V., Mittal, S., Maheshwari, S., and Balasubramanian, V.~N.
\newblock Memory-integrated reconfigurable adapters: A unified framework for settings with multiple tasks.
\newblock \emph{arXiv preprint arXiv:2512.00940}, 2025.

\bibitem[Anonymous(2026)]{anonymous2026remem}
Anonymous.
\newblock {REM}em: Reasoning with episodic memory in language agent, 2026.
\newblock Submitted to ICLR 2026.

\bibitem[Buckland(1991)]{buckland1991information}
Buckland, M.~K.
\newblock Information as thing.
\newblock \emph{Journal of the American Society for information science}, 42\penalty0 (5):\penalty0 351--360, 1991.

\bibitem[Cellier et~al.(2022)Cellier, Petersen, and Hwang]{cellier2022dynamics}
Cellier, D., Petersen, I.~T., and Hwang, K.
\newblock Dynamics of hierarchical task representations.
\newblock \emph{Journal of Neuroscience}, 42\penalty0 (38):\penalty0 7276--7284, 2022.

\bibitem[Chen et~al.(2026)Chen, Wu, Liu, and Zhang]{CHEN2026104336}
Chen, J., Wu, M., Liu, Q., and Zhang, Y.
\newblock Explainable prediction of knowledge recombination: A synergized method with heterogeneous hypergraph learning and large language models.
\newblock \emph{Information Processing \& Management}, 63\penalty0 (1):\penalty0 104336, 2026.
\newblock ISSN 0306-4573.
\newblock \doi{https://doi.org/10.1016/j.ipm.2025.104336}.
\newblock URL \url{https://www.sciencedirect.com/science/article/pii/S0306457325002778}.

\bibitem[Chhikara et~al.(2025)Chhikara, Khant, Aryan, Singh, and Yadav]{chhikara2025mem0}
Chhikara, P., Khant, D., Aryan, S., Singh, T., and Yadav, D.
\newblock Mem0: Building production-ready ai agents with scalable long-term memory.
\newblock \emph{arXiv preprint arXiv:2504.19413}, 2025.

\bibitem[Collins \& Quillian(1969)Collins and Quillian]{collins1969retrieval}
Collins, A.~M. and Quillian, M.~R.
\newblock Retrieval time from semantic memory.
\newblock \emph{Journal of verbal learning and verbal behavior}, 8\penalty0 (2):\penalty0 240--247, 1969.

\bibitem[Cormen et~al.(2022)Cormen, Leiserson, Rivest, and Stein]{cormen2022introduction}
Cormen, T.~H., Leiserson, C.~E., Rivest, R.~L., and Stein, C.
\newblock \emph{Introduction to algorithms}.
\newblock MIT press, 2022.

\bibitem[Cowan(2001)]{cowan2001magical}
Cowan, N.
\newblock The magical number 4 in short-term memory: A reconsideration of mental storage capacity.
\newblock \emph{Behavioral and brain sciences}, 24\penalty0 (1):\penalty0 87--114, 2001.

\bibitem[Cowan(2010)]{cowan2010magical}
Cowan, N.
\newblock The magical mystery four: How is working memory capacity limited, and why?
\newblock \emph{Current directions in psychological science}, 19\penalty0 (1):\penalty0 51--57, 2010.

\bibitem[Du et~al.(2025)Du, Li, Zhang, and Song]{du2025memr}
Du, X., Li, L., Zhang, D., and Song, L.
\newblock Memr$^{3}$: Memory retrieval via reflective reasoning for llm agents.
\newblock \emph{arXiv preprint arXiv:2512.20237}, 2025.

\bibitem[Fang et~al.(2025)Fang, Deng, Xu, Jiang, Tang, Xu, Deng, Yao, Wang, Qiao, et~al.]{fang2025lightmem}
Fang, J., Deng, X., Xu, H., Jiang, Z., Tang, Y., Xu, Z., Deng, S., Yao, Y., Wang, M., Qiao, S., et~al.
\newblock Lightmem: Lightweight and efficient memory-augmented generation.
\newblock \emph{arXiv preprint arXiv:2510.18866}, 2025.

\bibitem[Guti{\'e}rrez et~al.(2025)Guti{\'e}rrez, Shu, Qi, Zhou, and Su]{gutierrezrag}
Guti{\'e}rrez, B.~J., Shu, Y., Qi, W., Zhou, S., and Su, Y.
\newblock From rag to memory: Non-parametric continual learning for large language models.
\newblock In \emph{Forty-second International Conference on Machine Learning}, 2025.

\bibitem[Hopcroft et~al.(1983)Hopcroft, Ullman, and Aho]{hopcroft1983data}
Hopcroft, J.~E., Ullman, J.~D., and Aho, A.~V.
\newblock \emph{Data structures and algorithms}, volume 175.
\newblock Addison-wesley Boston, MA, USA:, 1983.

\bibitem[Jia et~al.(2025)Jia, Wu, Ding, Lu, and Zhang]{jia-etal-2025-hetgcot}
Jia, R., Wu, M., Ding, Y., Lu, J., and Zhang, Y.
\newblock {H}et{GC}o{T}: Heterogeneous graph-enhanced chain-of-thought {LLM} reasoning for academic question answering.
\newblock In Christodoulopoulos, C., Chakraborty, T., Rose, C., and Peng, V. (eds.), \emph{Findings of the Association for Computational Linguistics: EMNLP 2025}, pp.\  15950--15963, Suzhou, China, November 2025. Association for Computational Linguistics.
\newblock ISBN 979-8-89176-335-7.
\newblock \doi{10.18653/v1/2025.findings-emnlp.864}.
\newblock URL \url{https://aclanthology.org/2025.findings-emnlp.864/}.

\bibitem[Jiang et~al.(2025)Jiang, Hao, Cho, Li, Yuan, Chen, Ungar, Taylor, and Roth]{jiang2025know}
Jiang, B., Hao, Z., Cho, Y.-M., Li, B., Yuan, Y., Chen, S., Ungar, L., Taylor, C.~J., and Roth, D.
\newblock Know me, respond to me: Benchmarking llms for dynamic user profiling and personalized responses at scale.
\newblock \emph{arXiv preprint arXiv:2504.14225}, 2025.

\bibitem[Jiang et~al.()Jiang, Yang, Zhang, and Liu]{jiang2025towards}
Jiang, J., Yang, P., Zhang, R., and Liu, F.
\newblock Towards efficient large language model serving: A survey on system-aware kv cache optimization.
\newblock \emph{Authorea Preprints}.

\bibitem[Jonelagadda et~al.(2025)Jonelagadda, Hahn, Zheng, and Penachio]{jonelagadda2025mnemosyne}
Jonelagadda, A., Hahn, C., Zheng, H., and Penachio, S.
\newblock Mnemosyne: An unsupervised, human-inspired long-term memory architecture for edge-based llms.
\newblock \emph{arXiv preprint arXiv:2510.08601}, 2025.

\bibitem[Kang et~al.(2025)Kang, Ji, Zhao, and Bai]{kang2025memory}
Kang, J., Ji, M., Zhao, Z., and Bai, T.
\newblock Memory os of ai agent.
\newblock \emph{arXiv preprint arXiv:2506.06326}, 2025.

\bibitem[LangChain(2025)]{langchain2025langmem}
LangChain.
\newblock langmem: Language model memory.
\newblock \url{https://github.com/langchain-ai/langmem}, 2025.

\bibitem[Lewis et~al.(2020)Lewis, Perez, Piktus, Petroni, Karpukhin, Goyal, K{\"u}ttler, Lewis, Yih, Rockt{\"a}schel, et~al.]{lewis2020retrieval}
Lewis, P., Perez, E., Piktus, A., Petroni, F., Karpukhin, V., Goyal, N., K{\"u}ttler, H., Lewis, M., Yih, W.-t., Rockt{\"a}schel, T., et~al.
\newblock Retrieval-augmented generation for knowledge-intensive nlp tasks.
\newblock \emph{Advances in neural information processing systems}, 33:\penalty0 9459--9474, 2020.

\bibitem[Li et~al.(2025{\natexlab{a}})Li, Cai, Wang, Yu, and Chen]{li2025survey}
Li, X., Cai, Z., Wang, S., Yu, K., and Chen, F.
\newblock A survey on enhancing causal reasoning ability of large language models.
\newblock In \emph{Pacific-Asia Conference on Knowledge Discovery and Data Mining}, pp.\  399--416. Springer, 2025{\natexlab{a}}.

\bibitem[Li et~al.(2024)Li, Wen, Wang, Li, Yuan, Liu, Liu, Xu, Wang, Sun, et~al.]{li2024personal}
Li, Y., Wen, H., Wang, W., Li, X., Yuan, Y., Liu, G., Liu, J., Xu, W., Wang, X., Sun, Y., et~al.
\newblock Personal llm agents: Insights and survey about the capability, efficiency and security.
\newblock \emph{arXiv preprint arXiv:2401.05459}, 2024.

\bibitem[Li et~al.(2025{\natexlab{b}})Li, Xi, Li, Chen, Chen, Song, Niu, Wang, Yang, Tang, et~al.]{li2025memos}
Li, Z., Xi, C., Li, C., Chen, D., Chen, B., Song, S., Niu, S., Wang, H., Yang, J., Tang, C., et~al.
\newblock Memos: A memory os for ai system.
\newblock \emph{arXiv preprint arXiv:2507.03724}, 2025{\natexlab{b}}.

\bibitem[Lu et~al.(2025)Lu, Wu, Xu, Li, Liu, Ding, Sun, Lu, and Zhang]{lu2025newbornimpactbiasawarecitation}
Lu, M., Wu, M., Xu, J., Li, W., Liu, F., Ding, Y., Sun, Y., Lu, J., and Zhang, Y.
\newblock From newborn to impact: Bias-aware citation prediction, 2025.
\newblock URL \url{https://arxiv.org/abs/2510.19246}.

\bibitem[Maharana et~al.(2024)Maharana, Lee, Tulyakov, Bansal, Barbieri, and Fang]{maharana2024evaluating}
Maharana, A., Lee, D.-H., Tulyakov, S., Bansal, M., Barbieri, F., and Fang, Y.
\newblock Evaluating very long-term conversational memory of llm agents.
\newblock In \emph{Proceedings of the 62nd Annual Meeting of the Association for Computational Linguistics (Volume 1: Long Papers)}, pp.\  13851--13870, 2024.

\bibitem[Mathy \& Feldman(2012)Mathy and Feldman]{mathy2012s}
Mathy, F. and Feldman, J.
\newblock What’s magic about magic numbers? chunking and data compression in short-term memory.
\newblock \emph{Cognition}, 122\penalty0 (3):\penalty0 346--362, 2012.

\bibitem[Pan et~al.(2025)Pan, Wu, Jiang, Luo, Cheng, Li, Yang, Lin, Zhao, Qiu, and Gao]{pan2025secom}
Pan, Z., Wu, Q., Jiang, H., Luo, X., Cheng, H., Li, D., Yang, Y., Lin, C.-Y., Zhao, H.~V., Qiu, L., and Gao, J.
\newblock Secom: On memory construction and retrieval for personalized conversational agents.
\newblock In \emph{The Thirteenth International Conference on Learning Representations}, 2025.
\newblock URL \url{https://openreview.net/forum?id=xKDZAW0He3}.

\bibitem[Rasmussen et~al.(2025)Rasmussen, Paliychuk, Beauvais, Ryan, and Chalef]{rasmussen2025zep}
Rasmussen, P., Paliychuk, P., Beauvais, T., Ryan, J., and Chalef, D.
\newblock Zep: a temporal knowledge graph architecture for agent memory.
\newblock \emph{arXiv preprint arXiv:2501.13956}, 2025.

\bibitem[Schlegel et~al.(2025)Schlegel, Sommer, and Mortillaro]{schlegel2025large}
Schlegel, K., Sommer, N.~R., and Mortillaro, M.
\newblock Large language models are proficient in solving and creating emotional intelligence tests.
\newblock \emph{Communications Psychology}, 3\penalty0 (1):\penalty0 80, 2025.

\bibitem[Simon(2012)]{simon2012architecture}
Simon, H.~A.
\newblock The architecture of complexity.
\newblock In \emph{The Roots of Logistics}, pp.\  335--361. Springer, 2012.

\bibitem[Trach \& McDougle(2025)Trach and McDougle]{trach2025mental}
Trach, J.~E. and McDougle, S.~D.
\newblock Mental graphs structure the storage and retrieval of visuomotor associations.
\newblock \emph{Nature Human Behaviour}, pp.\  1--15, 2025.

\bibitem[Vogel \& Machizawa(2004)Vogel and Machizawa]{vogel2004neural}
Vogel, E.~K. and Machizawa, M.~G.
\newblock Neural activity predicts individual differences in visual working memory capacity.
\newblock \emph{Nature}, 428\penalty0 (6984):\penalty0 748--751, 2004.

\bibitem[Wang et~al.(2024)Wang, Ma, Feng, Zhang, Yang, Zhang, Chen, Tang, Chen, Lin, et~al.]{wang2024survey}
Wang, L., Ma, C., Feng, X., Zhang, Z., Yang, H., Zhang, J., Chen, Z., Tang, J., Chen, X., Lin, Y., et~al.
\newblock A survey on large language model based autonomous agents.
\newblock \emph{Frontiers of Computer Science}, 18\penalty0 (6):\penalty0 186345, 2024.

\bibitem[Wang et~al.(2025)Wang, Tian, Li, Liang, Wang, Chen, Wang, Lu, Ma, Jiang, et~al.]{wang2025mem}
Wang, P., Tian, M., Li, J., Liang, Y., Wang, Y., Chen, Q., Wang, T., Lu, Z., Ma, J., Jiang, Y.~E., et~al.
\newblock O-mem: Omni memory system for personalized, long horizon, self-evolving agents.
\newblock \emph{arXiv e-prints}, pp.\  arXiv--2511, 2025.

\bibitem[Wang \& Chen(2025)Wang and Chen]{wang2025mirix}
Wang, Y. and Chen, X.
\newblock Mirix: Multi-agent memory system for llm-based agents.
\newblock \emph{arXiv preprint arXiv:2507.07957}, 2025.

\bibitem[Wu et~al.(2025)Wu, Zhang, Haunschild, and Bornmann]{wu2025leveraging}
Wu, M., Zhang, Y., Haunschild, R., and Bornmann, L.
\newblock Leveraging large language models for post-publication peer review: Potential and limitations.
\newblock In \emph{Proceedings of the 20th International Conference on Scientometrics \& Informetrics (ISSI 2025)}, pp.\  1207--1226, 2025.

\bibitem[Xu et~al.(2025{\natexlab{a}})Xu, Wen, Jia, Zhang, wenlin zhang, Wang, Guo, Tang, Zhao, Chen, and Xu]{xu2025single}
Xu, D., Wen, Y., Jia, P., Zhang, Y., wenlin zhang, Wang, Y., Guo, H., Tang, R., Zhao, X., Chen, E., and Xu, T.
\newblock From single to multi-granularity: Toward long-term memory association and selection of conversational agents, 2025{\natexlab{a}}.
\newblock URL \url{https://arxiv.org/abs/2505.19549}.

\bibitem[Xu et~al.(2022)Xu, Szlam, and Weston]{xu2022beyond}
Xu, J., Szlam, A., and Weston, J.
\newblock Beyond goldfish memory: Long-term open-domain conversation.
\newblock In \emph{Proceedings of the 60th annual meeting of the association for computational linguistics (volume 1: long papers)}, pp.\  5180--5197, 2022.

\bibitem[Xu et~al.(2025{\natexlab{b}})Xu, Liang, Mei, Gao, Tan, and Zhang]{xu2025mem}
Xu, W., Liang, Z., Mei, K., Gao, H., Tan, J., and Zhang, Y.
\newblock A-mem: Agentic memory for llm agents.
\newblock \emph{arXiv preprint arXiv:2502.12110}, 2025{\natexlab{b}}.

\bibitem[Yan et~al.(2025)Yan, Yang, Huang, Nie, Ding, Li, Ma, Kersting, Pan, Sch{\"u}tze, et~al.]{yan2025memory}
Yan, S., Yang, X., Huang, Z., Nie, E., Ding, Z., Li, Z., Ma, X., Kersting, K., Pan, J.~Z., Sch{\"u}tze, H., et~al.
\newblock Memory-r1: Enhancing large language model agents to manage and utilize memories via reinforcement learning.
\newblock \emph{arXiv preprint arXiv:2508.19828}, 2025.

\bibitem[Yang et~al.(2025)Yang, Lu, and Yu]{yangadapting}
Yang, X., Lu, J., and Yu, E.
\newblock Adapting multi-modal large language model to concept drift from pre-training onwards.
\newblock In \emph{The Thirteenth International Conference on Learning Representations}, 2025.
\newblock URL \url{https://openreview.net/forum?id=b20VK2GnSs}.

\bibitem[Zhang et~al.(2025{\natexlab{a}})Zhang, Zhang, Zhang, Yang, Shang, Wei, Zou, Huang, Wang, Gao, Pan, Xiong, Liu, Yu, and Li]{zhang2025personaagentlargelanguagemodel}
Zhang, W., Zhang, X., Zhang, C., Yang, L., Shang, J., Wei, Z., Zou, H.~P., Huang, Z., Wang, Z., Gao, Y., Pan, X., Xiong, L., Liu, J., Yu, P.~S., and Li, X.
\newblock Personaagent: When large language model agents meet personalization at test time, 2025{\natexlab{a}}.
\newblock URL \url{https://arxiv.org/abs/2506.06254}.

\bibitem[Zhang et~al.(2025{\natexlab{b}})Zhang, Dai, Bo, Ma, Li, Chen, Zhu, Dong, and Wen]{zhang2025survey}
Zhang, Z., Dai, Q., Bo, X., Ma, C., Li, R., Chen, X., Zhu, J., Dong, Z., and Wen, J.-R.
\newblock A survey on the memory mechanism of large language model-based agents.
\newblock \emph{ACM Transactions on Information Systems}, 43\penalty0 (6):\penalty0 1--47, 2025{\natexlab{b}}.

\bibitem[Zhong et~al.(2024)Zhong, Guo, Gao, Ye, and Wang]{zhong2024memorybank}
Zhong, W., Guo, L., Gao, Q., Ye, H., and Wang, Y.
\newblock Memorybank: Enhancing large language models with long-term memory.
\newblock In \emph{Proceedings of the AAAI Conference on Artificial Intelligence}, volume~38, pp.\  19724--19731, 2024.

\end{thebibliography}
\bibliographystyle{icml2026}

\newpage
\appendix

\onecolumn
\section*{}
\label{sec:tech_appendix}

\vspace{0.5em}
\begin{center}
{\scshape Technical Appendix}
\end{center}
\vspace{0.8em}

\newcommand{\apptoc}[4]{%
  \noindent\textbf{#1}\quad
  \hyperref[#4]{#2}\dotfill #3\par
}

\newcommand{\apptocsub}[4]{%
  \noindent\hspace*{1.5em}\textbf{#1}\quad
  \hyperref[#4]{#2}\dotfill #3\par
}

\apptoc{A}{Description of Datasets, Baselines and Evaluation Metrics}{10}{sec:appendix_datasets_baselines}
\apptocsub{A.1}{Datasets}{11}{sec:datasets}
\apptocsub{A.2}{Baselines}{12}{sec:baselines}
\apptocsub{A.3}{Evaluation Metrics}{13}{sec:eval_metrics}

\apptoc{B}{Beta-Mixture Gating Model (BMM)}{14}{app:bmm}

\apptoc{C}{Conversation Feature Definition for Structure Selection}{15}{sec:conversation_features}
\apptocsub{C.1}{Feature Overview}{15}{sec:conversation_features}
\apptocsub{C.2}{Feature Definition Table}{15}{sec:conversation_features}

\apptoc{D}{Structure Selector Training Pipeline}{16}{sec:structure_selector_training}

\apptoc{E}{Additional experiments results}{18}{sec:additional_results}

\apptoc{F}{Memory Structure Implementations}{18}{sec:memorystructure}

\apptoc{G}{Computational Cost Analysis}{18}{sec:computational cost analysis}

\apptoc{H}{Case Studies}{19}{sec:case}

\apptoc{I}{LLM Prompts Design}{22}{sec:llm_prompts}
\apptocsub{I.1}{Response}{22}{fig3}
\apptocsub{I.2}{Meta Info}{22}{fig5}
\apptocsub{I.3}{Entity Relation Extraction}{23}{fig6}
\apptocsub{I.4}{Hierarchical Structure Construction}{24}{fig7}
\apptocsub{I.5}{Procedural Extraction}{24}{fig7}

\section{Description of Datasets, Baselines and Evaluation Metrics}
\label{sec:appendix_datasets_baselines}

\subsection{Datasets}
\label{sec:datasets}
\textbf{LoCoMo}~\citep{maharana2024evaluating} 
is a benchmark designed to evaluate LLM agents' long-term conversational memory. It consists of very long dialogue traces, where each conversation contains on average $\sim$300 turns and $\sim$9{,}000 tokens, spanning months of interactions (typically 19--35 sessions). LoCoMo operationalizes memory and reasoning evaluation by grouping questions into five categories:
\begin{itemize}
  \item \textbf{Single-hop reasoning:} the answer can be retrieved from information in a single session.
  \item \textbf{Multi-hop reasoning:} the answer requires aggregating evidence scattered across multiple sessions.
  \item \textbf{Temporal reasoning:} tests the ability to track temporal cues and reason over time gaps in the dialogue history.
  \item \textbf{Open-domain:} requires combining user-provided facts with external world knowledge or commonsense.
  \item \textbf{Adversarial:} intentionally crafted to trigger hallucinations; the desired behavior is to recognize the question as unanswerable.
\end{itemize}

\textbf{PERSONAMEM}~\citep{jiang2025know} 
is a benchmark designed to evaluate LLMs’ ability to perform \emph{dynamic user profiling} and generate personalized responses over long-horizon interactions.
It consists of multi-session dialogue histories in which user attributes and preferences evolve over time due to life events or contextual changes.
PERSONAMEM operationalizes personalization and memory evaluation by grouping in-situ user queries into seven categories:

\begin{itemize}
  \item \textbf{Recall user-shared facts:} tests whether the model can retrieve static user information explicitly mentioned in earlier interactions.
  \item \textbf{Acknowledge latest preferences:} evaluates the ability to identify and use the user’s most recent preference state when older information becomes outdated.
  \item \textbf{Track preference evolution:} requires reconstructing how a user’s preferences change across multiple sessions.
  \item \textbf{Revisit reasons behind preference updates:} tests whether the model can recall the underlying events or causes that led to preference changes.
  \item \textbf{Suggest new ideas:} asks the model to propose novel suggestions consistent with the user’s inferred preferences.
  \item \textbf{Provide preference-aligned recommendations:} evaluates personalized recommendation quality conditioned on the current user profile.
  \item \textbf{Generalize to new scenarios:} tests whether user-specific knowledge learned in one domain can be transferred to unseen tasks or contexts.
\end{itemize}

\subsection{Baselines}
\label{sec:baselines}
\textbf{Mem0}~\citep{chhikara2025mem0} is a scalable long-term memory architecture specifically designed for AI agents to address the challenge that large language models cannot maintain consistency in prolonged multi-turn conversations due to fixed context window limitations. This method comprises two core stages: "extraction" and "update". First, it leverages LLMs combined with context to extract key facts from new conversations. Subsequently, through intelligent decision-making (Tool Call), it executes ADD, UPDATE, DELETE, or NOOP instructions to dynamically maintain memory consistency. Additionally, Mem0 provides an enhanced variant ($Mem0^s$) that integrates graph databases by using entity nodes and relationship edges to capture complex information structures.

\textbf{LangMem}~\citep{langchain2025langmem} is a long-term memory management library developed by the LangChain team to help AI agents learn and adapt from interactions. It provides tools for extracting important information from conversations, optimizing agent behavior, and maintaining long-term memory, with support for arbitrary storage systems and native LangGraph integration. Core features include: real-time memory management tools that enable agents to actively record and search information, background memory managers that automatically extract and integrate knowledge, and seamless integration with the LangGraph platform, enabling agents to continuously improve and personalize responses across sessions. 

\textbf{MemoryOS}~\citep{kang2025memory} is an OS-inspired memory management framework that optimizes long-term memory for LLM agents through a multi-tiered architecture. It includes short-term, mid-term, and long-term memory layers, allowing dynamic updates and integration during conversations. The framework features memory storage, update, retrieval, and generation modules, facilitating efficient memory decay and redundancy filtering. By applying OS principles, MemoryOS enhances response consistency and coherence while minimizing computational overhead.

\textbf{A-MEM}~\citep{xu2025mem} is an agentic memory architecture that organizes memory as a network of interconnected notes. It employs dedicated agents to manage note creation, linking, and retrieval, enabling dynamic memory evolution through reflection and integration of new information. A-MEM supports hierarchical and relational reasoning by maintaining structured connections between notes, facilitating effective recall during interactions.

\textbf{ZEP}~\citep{rasmussen2025zep} is a time-aware graph memory architecture designed for LLM agents to maintain consistency across multi-session interactions. It constructs a knowledge graph that captures temporal relationships between memory entries, allowing agents to reason about changes over time. ZEP employs specialized retrieval mechanisms to access relevant information based on temporal context, enhancing the agent's ability to provide coherent and contextually appropriate responses.

\textbf{MEMOS}~\citep{li2025memos} is a memory operating system designed to treat memory as a schedulable and evolvable system resource for Large Language Models (LLMs). It establishes a unified hierarchy that integrates three distinct memory types—plaintext, activation, and parameter memory—encapsulated within a standardized unit called the \textit{MemCube}. By employing a layered architecture featuring a scheduler and lifecycle manager, MEMOS enables the dynamic transformation, storage, and retrieval of memory, thereby facilitating long-term consistency, adaptive personalization, and efficient knowledge evolution across complex tasks.

\textbf{MemR$^3$}~\citep{du2025memr} is an autonomous memory retrieval system designed to transform the standard retrieve-then-answer pipeline into a closed-loop sequential decision process for LLM agents. Implemented as a LangGraph workflow, it utilizes a router to dynamically select actions among `retrieve`, `reflect`, and `answer` to optimize response quality. Central to its architecture is a global evidence-gap tracker that explicitly maintains verified evidence and identifies missing information gaps throughout the interaction. This design enables the system to iteratively refine queries and stop retrieval early once sufficient information is gathered, functioning as a plug-and-play controller compatible with various existing memory backends.

\textbf{O-Mem}~\citep{wang2025mem} is an omni-memory framework designed to enhance the long-term interaction and dynamic personalization of LLM agents. Unlike traditional systems that passively store interaction history, O-Mem utilizes an \textit{active user profiling} mechanism that dynamically extracts and updates user characteristics and event records during ongoing dialogues. It employs a hierarchical retrieval strategy across three distinct components—Persona Memory (for long-term attributes), Working Memory (for topical context), and Episodic Memory (for clue-triggered recall)—to construct a holistic user model. This architecture enables the agent to maintain contextual consistency over long horizons and self-evolve its understanding of the user, leading to more adaptive and coherent personalized responses.

\textbf{HippoRAG}~\citep{gutierrezrag} is a neurobiologically inspired framework that repurposes Retrieval-Augmented Generation (RAG) as a non-parametric continual learning system for LLMs. Grounded in the hippocampal indexing theory, it maintains a dynamic Knowledge Graph (KG) that continuously integrates new information through simulated synaptic plasticity, rather than static vector indexing. By employing the Personalized PageRank (PPR) algorithm for retrieval, HippoRAG enables pattern completion and complex multi-hop reasoning across temporally distributed facts, allowing the model to adapt to new knowledge in a single pass without parametric training or catastrophic forgetting.

\subsection{Evaluation Metrics}
\label{sec:eval_metrics}

\textbf{F1}: 
The F1 score is computed as the harmonic mean of precision and recall:
\begin{equation}
F1 = 2 \cdot \frac{\text{Precision} \times \text{Recall}}{\text{Precision} + \text{Recall}},
\end{equation}
where Precision $= \frac{TP}{TP + FP}$ and Recall $= \frac{TP}{TP + FN}$, with $TP$, $FP$, and $FN$ denoting true positives, false positives, and false negatives, respectively.

\textbf{Bleu-1}:  
BLEU (Bilingual Evaluation Understudy) is a metric for evaluating machine translation and text generation quality. BLEU-1 measures the proportion of unigrams (single words) in the generated text that appear in the reference text:
\begin{equation}
    \text{BLEU-1} = BP \cdot \exp \left( \sum_{n=1}^{1} w_n \log p_n \right),
\end{equation}

where
\begin{equation}
    BP = \begin{cases} 
    1 & \text{if } c > r \\ 
    e^{1-r/c} & \text{if } c \le r
    \end{cases} ,
\end{equation}

\begin{equation}
    p_n = \frac{\sum_{i} \sum_{k} \min(h_{ik}, m_{ik})}{\sum_{i} \sum_{k} h_{ik}},
\end{equation}
Here, $BP$ denotes the brevity penalty used to penalize short system outputs, where $c$ is the length of the candidate translation and $r$ is the effective reference length. The term $w_n$ represents the weight assigned to $n$-gram precision $p_n$. Specifically, $h_{ik}$ is the count of the $k$-th $n$-gram in the $i$-th candidate sentence, while $m_{ik}$ is its maximum frequency in the corresponding reference. The $\min(h_{ik}, m_{ik})$ function ensures the counts are clipped to prevent overestimation of precision from repetitive words.

\section{Beta-Mixture Gating Model (BMM)}
\label{app:bmm}

\paragraph{Score normalization.}
Given an incoming memory item, we compute matching scores against $n$ candidate sessions, producing
$\{s_i\}_{i=1}^{n}$.
Since Beta distributions are defined on $(0,1)$, we apply min--max normalization:
\begin{equation}
x_i =
\begin{cases}
\epsilon + (1-2\epsilon)\dfrac{s_i - s_{\min}}{s_{\max}-s_{\min}}, & s_{\max} > s_{\min},\\[6pt]
0.5, & s_{\max} = s_{\min},
\end{cases}
\qquad x_i\in(0,1),
\end{equation}
where $s_{\min}=\min_i s_i$, $s_{\max}=\max_i s_i$, and $\epsilon$ is a small constant to avoid boundary values.

\paragraph{Two-component Beta mixture.}
We fit a two-component Beta mixture model to $\{x_i\}$:
\begin{equation}
p(x)=\pi_0\,\mathrm{Beta}(x;\alpha_0,\beta_0)+\pi_1\,\mathrm{Beta}(x;\alpha_1,\beta_1),
\qquad \pi_k\ge 0,\ \sum_{k\in\{0,1\}}\pi_k=1,
\end{equation}
with
\begin{equation}
\mathrm{Beta}(x;\alpha,\beta)=\frac{\Gamma(\alpha+\beta)}{\Gamma(\alpha)\Gamma(\beta)}x^{\alpha-1}(1-x)^{\beta-1}.
\end{equation}
We interpret the two components as \emph{low-compatibility} vs.\ \emph{high-compatibility} regimes.

\paragraph{EM fitting with log-space responsibilities.}
We estimate parameters via EM for a fixed number of iterations (or until convergence).
Introduce latent assignments $z_i\in\{0,1\}$ and compute responsibilities
$r_{ik}=p(z_i=k\mid x_i)$.

\subparagraph{Initialization (quantile-based).}
We initialize the two components to be separated in mean by using the empirical 30th and 70th percentiles of $\{x_i\}$.
Let $q_{0.3}$ and $q_{0.7}$ denote these quantiles.
We initialize component means as $\mu_0\leftarrow q_{0.3}$ and $\mu_1\leftarrow q_{0.7}$ and set initial concentrations
(e.g., via a small fixed variance), then obtain $(\alpha_k,\beta_k)$ by moment matching.

\subparagraph{E-step.}
For numerical stability, we compute responsibilities in log-space:
\begin{equation}
\log \tilde{r}_{ik}
=
\log \pi_k + \log \mathrm{Beta}(x_i;\alpha_k,\beta_k),
\qquad
r_{ik} = \frac{\exp(\log \tilde{r}_{ik})}{\sum_{k'\in\{0,1\}}\exp(\log \tilde{r}_{ik'})}.
\end{equation}
Define effective counts $N_k=\sum_{i=1}^{n} r_{ik}$.

\subparagraph{M-step (mixture weights).}
\begin{equation}
\pi_k \leftarrow \frac{N_k}{n}.
\end{equation}


\begin{algorithm}[t]
\caption{Beta-Mixture-Gated Memory Fusion}
\label{alg:bmm_gate}
\begin{algorithmic}[1]
\REQUIRE incoming item $m$, candidates $\mathcal{C}$, threshold $\tau$, min-keep $m_{\min}$, EM iters $T$
\STATE $s_i \gets \mathrm{score}(m,c_i)$ for each $c_i\in\mathcal{C}$
\STATE $x_i \gets \mathrm{minmax\_to\_unit}(s_i)$
\IF{$\max s_i = \min s_i$} \STATE $x_i \gets 0.5$ for all $i$ \ENDIF
\STATE Initialize 2 components using quantiles $(q_{0.3}, q_{0.7})$ of $\{x_i\}$
\FOR{$t=1$ to $T$}
  \STATE \textbf{E-step:} $r_{ik} \propto \pi_k \mathrm{Beta}(x_i;\alpha_k,\beta_k)$ (computed in log-space)
  \STATE \textbf{M-step:} $\pi_k \gets \frac{1}{n}\sum_i r_{ik}$; update $(\alpha_k,\beta_k)$ by weighted moments
\ENDFOR
\STATE $k^\star \gets \arg\max_k \frac{\alpha_k}{\alpha_k+\beta_k}$
\STATE $g_i \gets p(z=k^\star \mid x_i)$ for all $i$
\STATE $\mathcal{I} \gets \{i: g_i \ge \tau\}$
\IF{$|\mathcal{I}| < m_{\min}$} \STATE $\mathcal{I} \gets \mathrm{TopK}(\{x_i\},m_{\min})$ \ENDIF
\IF{$\mathcal{I}=\emptyset$}
    \STATE \textbf{return} \textsc{NewSession}$(m)$
\ENDIF
\STATE $s^\star \gets \arg\max_{i\in\mathcal{I}} x_i$
\STATE \textbf{return} \textsc{Merge}$(m,c_{s^\star})$
\end{algorithmic}
\end{algorithm}

\subparagraph{M-step (Beta parameters via responsibility-weighted moments).}
Rather than numerically optimizing the Beta log-likelihood, we adopt a lightweight update using weighted moments.
Compute the responsibility-weighted mean and variance:
\begin{equation}
\mu_k=\frac{1}{N_k}\sum_{i=1}^{n} r_{ik}x_i,
\qquad
\sigma_k^2=\frac{1}{N_k}\sum_{i=1}^{n} r_{ik}(x_i-\mu_k)^2.
\end{equation}
For Beta$(\alpha,\beta)$, $\mu=\alpha/(\alpha+\beta)$ and
$\sigma^2=\frac{\mu(1-\mu)}{\kappa+1}$ with concentration $\kappa=\alpha+\beta$.
Thus,
\begin{equation}
\kappa_k \leftarrow \frac{\mu_k(1-\mu_k)}{\sigma_k^2}-1,
\qquad
\alpha_k \leftarrow \mu_k\kappa_k,
\qquad
\beta_k \leftarrow (1-\mu_k)\kappa_k.
\end{equation}
In practice, we clamp $\sigma_k^2$ away from zero and clip $\kappa_k,\alpha_k,\beta_k$ to avoid degenerate updates.

\paragraph{Identifying the high-compatibility component.}
After fitting, we designate the component with larger Beta mean as the \emph{high} component:
\begin{equation}
k^{\star}=\arg\max_{k\in\{0,1\}} \frac{\alpha_k}{\alpha_k+\beta_k}.
\end{equation}
The gate value for a candidate score $x$ is the posterior of this high component:
\begin{equation}
g(x)=p(z=k^{\star}\mid x)
=\frac{\pi_{k^{\star}}\mathrm{Beta}(x;\alpha_{k^{\star}},\beta_{k^{\star}})}
{\sum_{k\in\{0,1\}}\pi_{k}\mathrm{Beta}(x;\alpha_{k},\beta_{k})}.
\end{equation}

\paragraph{Thresholding with minimum-keep fallback.}
Given gate values $\{g(x_i)\}$, we retain candidates satisfying
\begin{equation}
g(x_i)\ge \tau,
\end{equation}
where $\tau$ is a posterior threshold.
To prevent over-filtering, if fewer than $m_{\min}$ candidates satisfy the condition, we keep the top-$m_{\min}$
candidates ranked by $x_i$ (equivalently, by the original matching scores).

\paragraph{Usage in memory fusion.}
The retained candidates define a filtered set of plausible merge targets.
The incoming memory item is merged into the most compatible retained session (under the currently selected structure),
or stored as a new session if no candidate is retained.
This procedure replaces fixed similarity thresholds with an adaptive, distribution-aware gating rule that is robust to
score distribution shifts across conversation types and interaction stages.

\paragraph{Computational complexity.}
For $n$ candidate sessions, each EM iteration runs in $\mathcal{O}(n)$ time.
Since the number of mixture components is fixed to two and the number of EM iterations is small and bounded,
the overall gating cost is negligible compared to retrieval and response generation.
The memory overhead is $\mathcal{O}(n)$ for storing normalized scores and responsibilities.

\section{Conversation Feature Definition for Structure Selection}
\label{sec:conversation_features}
\subsection{Feature Overview}
We extract a compact set of interpretable interaction-level features to characterize the structural properties of the ongoing conversation.
These features are designed to capture temporal dynamics, entity-centricity, and topic evolution patterns, which are indicative of suitable memory organizations.

\subsection{Feature Definition Table}

\begin{table*}[t]
\centering
\caption{Conversation-level features used for memory structure selection. All features are lightweight to compute and designed to capture structural characteristics rather than fine-grained semantics.}
\begin{tabular}{l p{0.65\linewidth}}
\toprule
\textbf{Feature} & \textbf{Description and Motivation} \\
\midrule
\texttt{page\_count} &
Number of interaction pages within the current context window.
Captures interaction scale; longer conversations often benefit from structured or hierarchical memory. \\

\texttt{avg\_page\_length} &
Average length of recent pages (in tokens or turns).
Longer pages indicate information-dense interactions that may require non-linear organization. \\

\texttt{entity\_density} &
Average number of named entities per page.
High values suggest entity-centric interactions, favoring graph-based memory. \\

\texttt{relation\_indicators} &
Frequency of explicit relational expressions (e.g., comparisons, references, dependencies).
Signals the presence of relational structure beyond pure temporal order. \\

\texttt{topic\_diversity} &
Number of distinct topics detected in recent interactions.
High diversity indicates branching or hierarchical discourse structure. \\

\texttt{topic\_transitions} &
Frequency of topic shifts across consecutive pages.
Rapid transitions favor hierarchical or graph-based organization over linear memory. \\

\texttt{is\_qna\_pattern} &
Binary indicator of question--answer dominated interaction patterns.
Such patterns often align with linear, recency-driven memory access. \\

\texttt{is\_decision\_tree} &
Binary indicator of decision-tree-like dialogue flow (e.g., successive conditional questions).
Suggests hierarchical organization with coarse-to-fine retrieval. \\

\texttt{is\_entity\_centric} &
Binary indicator of entity-focused queries (e.g., repeated mentions of the same entities).
Favors graph-based memory structures. \\

\texttt{time\_span} &
Temporal span covered by recent interactions.
Longer spans reduce the effectiveness of pure recency-based retrieval. \\

\texttt{temporal\_density} &
Interaction frequency per unit time.
Dense interactions emphasize short-term dependencies and linear access. \\

\texttt{semantic\_complexity} &
Estimated semantic variability across recent pages (e.g., embedding dispersion).
High complexity suggests heterogeneous content requiring structured organization. \\
\bottomrule
\end{tabular}
\label{tab:structure_features}
\end{table*}

As summarized in Table~\ref{tab:structure_features}, these features are intentionally simple and interpretable, focusing on structural signals rather than deep semantic representations.
This design keeps structure selection lightweight and stable, and avoids introducing additional reasoning overhead into the control layer.

\section{Structure Selector Training Pipeline}
\label{sec:structure_selector_training}
\paragraph{Offline supervision for structure selection.}
Directly optimizing memory structure selection via online reinforcement learning is undesirable in our setting, as it introduces instability, high variance, and additional interaction cost.
Instead, we adopt an offline supervision strategy that derives structure labels from interaction-level outcomes.
Specifically, we evaluate each candidate memory structure under identical interaction contexts and assign the structure that yields the highest empirical reward as the target label (Algorithm~\ref{alg:offline_label}).
This formulation reduces structure selection to a standard supervised classification problem, enabling stable training while grounding decisions in downstream response quality (Algorithm~\ref{alg:train_selector}).

\begin{algorithm}[H]
\caption{Offline Structure Labeling via Interaction-Derived Rewards}
\label{alg:offline_label}
\begin{algorithmic}[1]
\REQUIRE Training conversations $\mathcal{D}$, structures $\mathcal{S}=\{\textsc{Linear},\textsc{Graph},\textsc{Hierarchical}\}$
\REQUIRE Fixed agent pipeline $\mathcal{A}(\cdot;\, s)$ instantiated with structure $s$
\REQUIRE Reward weights $\lambda_{\text{judge}}, \lambda_{\text{mem}}$ (e.g., 0.7/0.3)
\ENSURE Labeled dataset $\mathcal{T}=\{(x, y)\}$

\STATE $\mathcal{T} \gets \emptyset$
\FOR{each conversation/session instance $\xi \in \mathcal{D}$}
    \STATE Extract feature vector $x \gets \mathrm{Feat}(\xi)$
    \FOR{each $s \in \mathcal{S}$}
        \STATE Run agent with structure $s$: $\hat{a}_s \gets \mathcal{A}(\xi;\, s)$
        \STATE Compute judge reward: $r_{\text{judge}}(s) \gets \mathrm{Judge}(\hat{a}_s, \xi)$
        \STATE Compute memory utilization reward: $r_{\text{mem}}(s) \gets \mathrm{MemUtil}(\hat{a}_s, \xi)$
        \STATE $r(s) \gets \lambda_{\text{judge}}\, r_{\text{judge}}(s) + \lambda_{\text{mem}}\, r_{\text{mem}}(s)$
    \ENDFOR
    \STATE $y \gets \arg\max_{s\in\mathcal{S}} r(s)$ \STATE {\footnotesize\textit{(reward-optimal structure label)}}
    \STATE $\mathcal{T} \gets \mathcal{T} \cup \{(x, y)\}$
\ENDFOR
\STATE \textbf{return} $\mathcal{T}$
\end{algorithmic}
\end{algorithm}

\begin{algorithm}[H]
\caption{Supervised Training of Structure Selector}
\label{alg:train_selector}
\begin{algorithmic}[1]
\REQUIRE Labeled dataset $\mathcal{T}=\{(x_i,y_i)\}$, selector $f_\theta(\cdot)$ (MLP)
\REQUIRE Loss $\mathcal{L}$ (cross-entropy), optimizer \textsc{Adam}
\ENSURE Trained parameters $\theta$

\STATE Fit feature normalizer (e.g., \textsc{StandardScaler}) on $\{x_i\}$
\FOR{$\text{epoch}=1$ to $\text{MaxEpoch}$}
    \FOR{each mini-batch $\mathcal{B}\subset\mathcal{T}$}
        \STATE $\tilde{x} \gets \mathrm{Normalize}(x)$ for all $(x,y)\in\mathcal{B}$
        \STATE $p \gets \mathrm{Softmax}(f_\theta(\tilde{x}))$
        \STATE $\theta \gets \theta - \eta \nabla_\theta \sum_{(x,y)\in\mathcal{B}}\mathcal{L}(p,y)$
    \ENDFOR
    \STATE \textit{Early stop if validation performance does not improve}
\ENDFOR
\STATE \textbf{return} $\theta$
\end{algorithmic}
\end{algorithm}

\newpage
\section{Additional Experimental Results}
\label{sec:additional_results}

Table~\ref{tab:locomo_appendix} reports category-wise results on \textsc{LoCoMo} using ROUGE-1, ROUGE-2, and BERTScore.
Overall, \textbf{FluxMem} achieves the strongest and most balanced performance across all four reasoning categories.
The improvements are particularly pronounced in multi-hop and single-hop reasoning, where FluxMem consistently attains the highest ROUGE and BERTScore values, indicating more accurate content selection and semantic alignment.
On open-domain questions, FluxMem substantially outperforms baselines in ROUGE-1, reflecting its ability to retrieve relevant information under weak structural cues.
While some baselines achieve competitive BERTScore in specific categories, they fail to maintain consistent performance across metrics and reasoning types.
These results further confirm that adaptive multi-structure memory organization enables more robust retrieval and generation across diverse long-horizon reasoning scenarios.

\begin{table*}[h]
  \centering
  \caption{Performance comparison on LoCoMo across different reasoning categories. Best results are shown in bold, second-best are underlined.}
  \small
  \setlength{\tabcolsep}{3.5pt}
  \begin{tabular}{l|ccc|ccc|ccc|ccc}
    \toprule
    \multirow{2}{*}{Method} 
    & \multicolumn{3}{c|}{Cat1: Multi-hop} 
    & \multicolumn{3}{c|}{Cat2: Temporal} 
    & \multicolumn{3}{c|}{Cat3: Open} 
    & \multicolumn{3}{c}{Cat4: Single-hop} \\
    & R-1 & R-2 & BS 
    & R-1 & R-2 & BS 
    & R-1 & R-2 & BS 
    & R-1 & R-2 & BS \\
    \midrule
    LangMem~\citep{langchain2025langmem} 
    & \underline{48.07} & \underline{13.80} & 65.16 
    & \textbf{64.12} & 21.55 & 80.87 
    & 24.55 & \underline{9.38} & 45.19 
    & \underline{51.30} & 29.15 & 68.49 \\

    Mem0~\citep{chhikara2025mem0}     
    & 29.95 & 9.43 & 52.20 
    & 24.31 & 6.94 & 50.27 
    & 23.53 & 5.56 & 40.82 
    & 33.48 & 19.25 & 46.42 \\

    ZEP~\citep{rasmussen2025zep} 
    & 8.00 & 2.11 & 32.38 
    & 6.56 & 1.06 & 19.05 
    & 6.10 & 0.92 & 34.09 
    & 6.94 & 2.26 & 26.29 \\

    MemoryOS~\citep{kang2025memory}  
    & 44.27 & 17.19 & \underline{89.91}
    & 44.87 & 19.84 & \underline{91.17}
    & \underline{28.20} & \textbf{9.68} & \underline{88.67}
    & 50.77 & \underline{30.29} & \underline{91.07} \\

    A-Mem~\citep{xu2025mem}     
    & 34.54 & 13.62 & 54.95 
    & 40.86 & 19.01 & 65.67 
    & 16.85 & 3.66 & 40.06 
    & 43.06 & 25.66 & 56.13 \\

    MEMOS~\citep{li2025memos} 
    & 18.33 & 6.50 & 50.75 
    & 12.40 & 3.25 & 44.05 
    & 11.04 & 2.45 & 39.91 
    & 18.10 & 10.01 & 53.30 \\

    MemR$^{3}$~\citep{du2025memr}   
    & 17.34 & 6.20 & -- 
    & 24.15 & 8.00 & -- 
    & 10.39 & 2.26 & -- 
    & 23.09 & 13.77 & -- \\

    O-mem~\citep{wang2025mem}    
    & 36.83 & 14.53 & 30.81 
    & 50.02 & \textbf{26.45} & 50.38 
    & 21.62 & 3.19 & 26.84 
    & 45.16 & 27.74 & 41.60 \\

    HippoRAG 2~\citep{gutierrezrag} 
    & 29.19 & 9.46 & 87.26 
    & 10.32 & 2.12 & 84.32 
    & 25.08 & 7.70 & 88.05
    & 45.70 & 29.23 & 90.24 \\

    \textbf{FluxMem} 
    & \textbf{49.80} & \textbf{19.34} & \textbf{90.39}
    & \underline{56.96} & \underline{23.95} & \textbf{92.81}
    & \textbf{37.42} & 9.29 & \textbf{89.47}
    & \textbf{63.83} & \textbf{38.98} & \textbf{92.58} \\
    \bottomrule
  \end{tabular}
  \label{tab:locomo_appendix}
\end{table*}

\section{Memory Structure Implementations}
\label{sec:memorystructure}
For completeness, we briefly describe the implementation details of the three memory structures considered in Sec. 3.3. These structures are not proposed in this work but are adapted from prior studies on long-term memory for LLM agents.

Linear Memory.
Linear memory organizes episodic sessions in strict chronological order. Each memory unit is appended sequentially, and retrieval is primarily driven by recency or similarity-based ranking. This structure favors temporal coherence and is effective when recent interactions dominate relevance.

Graph Memory.
Graph memory represents episodic sessions as nodes in a graph, where edges encode semantic or entity-level relations (e.g., shared entities or high embedding similarity). Retrieval is performed by traversing structurally related nodes rather than relying solely on temporal proximity, enabling multi-hop relational recall.

Hierarchical Memory.
Hierarchical memory clusters episodic sessions into higher-level abstractions (e.g., topics or summaries), forming a tree-like structure. Retrieval proceeds in a coarse-to-fine manner, first selecting  relevant high-level clusters and then refining to specific episodes, which improves scalability and abstraction.

\section{Computational Cost Analysis}
\label{sec:computational cost analysis}
We briefly analyze the computational overhead introduced by FluxMem. Overall, the additional cost mainly comes from memory selection and fusion, while the backbone LLM inference remains unchanged.

Memory structure selection.
The memory structure selector is a two-layer MLP with a 12-dimensional input and hidden size 4, incurring negligible overhead compared to LLM inference. This module is executed once per interaction and adds a constant-time cost.

BMM-based memory fusion.
The Beta Mixture Model (BMM) gate operates on matching scores of candidate memories and involves lightweight posterior computation. Since fusion is applied only to the retrieved MTEM candidates, its complexity scales linearly with the number of candidates and remains small in practice.

Retrieval cost.
Retrieval combines dense encoding and BM25 via reciprocal rank fusion. Dense retrieval relies on pre-computed embeddings, and BM25 operates over indexed text, making the overall retrieval cost comparable to standard hybrid retrieval pipelines commonly used in LLM-based systems.

Overall cost.
Compared to baselines with fixed memory structures, FluxMem introduces modest additional computation for structure selection and probabilistic gating. Empirically, this overhead is minor relative to LLM decoding cost and does not significantly affect end-to-end latency, while enabling more adaptive and selective memory utilization.

\section{Case Studies}
\label{sec:case}

\begin{figure*}[t]
  \centering
  \includegraphics[width=0.8\linewidth]{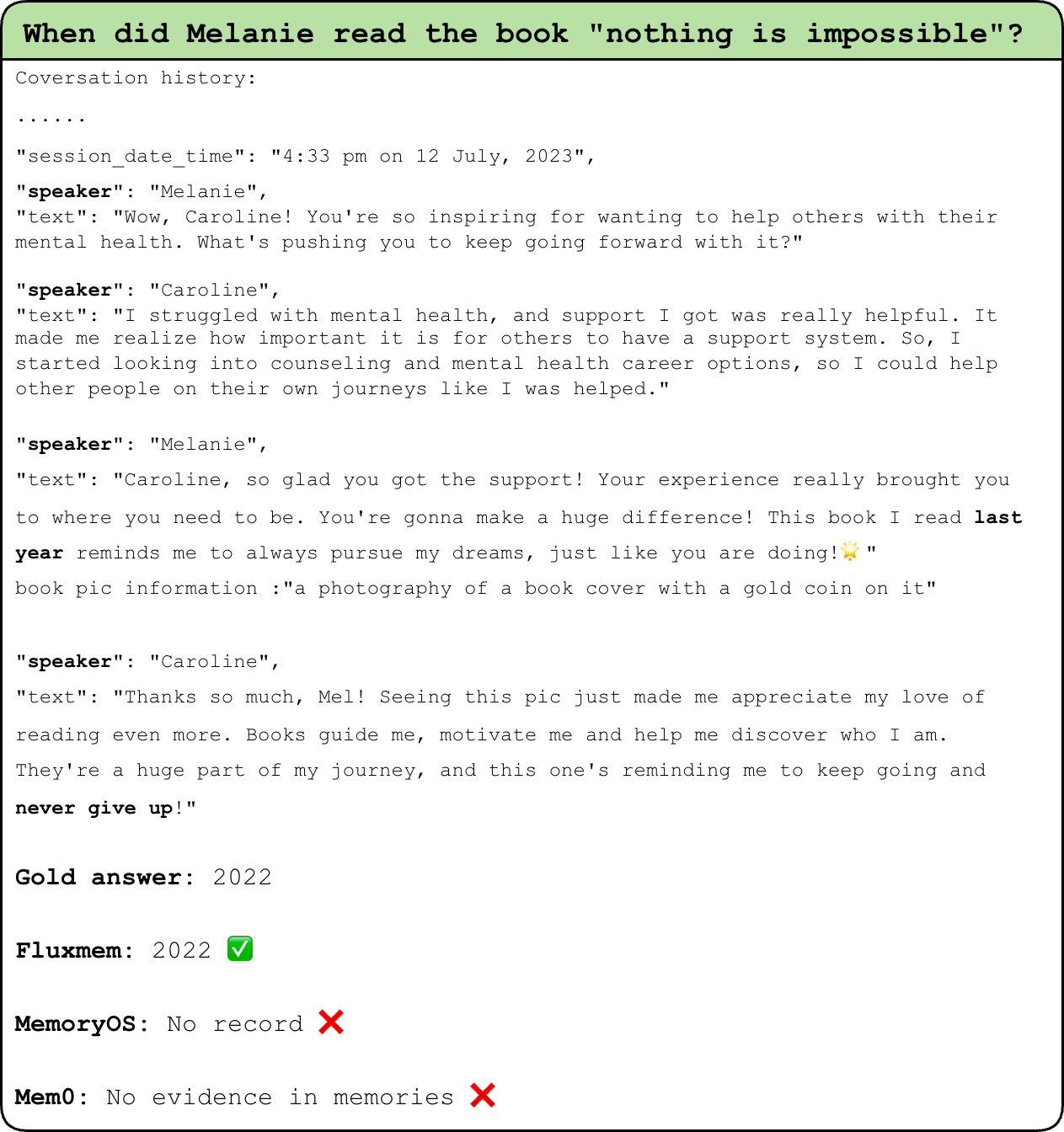}
  \caption{Time-dependent query case.}
  \label{figcase1}
\end{figure*}

\begin{figure*}[t]
  \centering
  \includegraphics[width=0.8\linewidth]{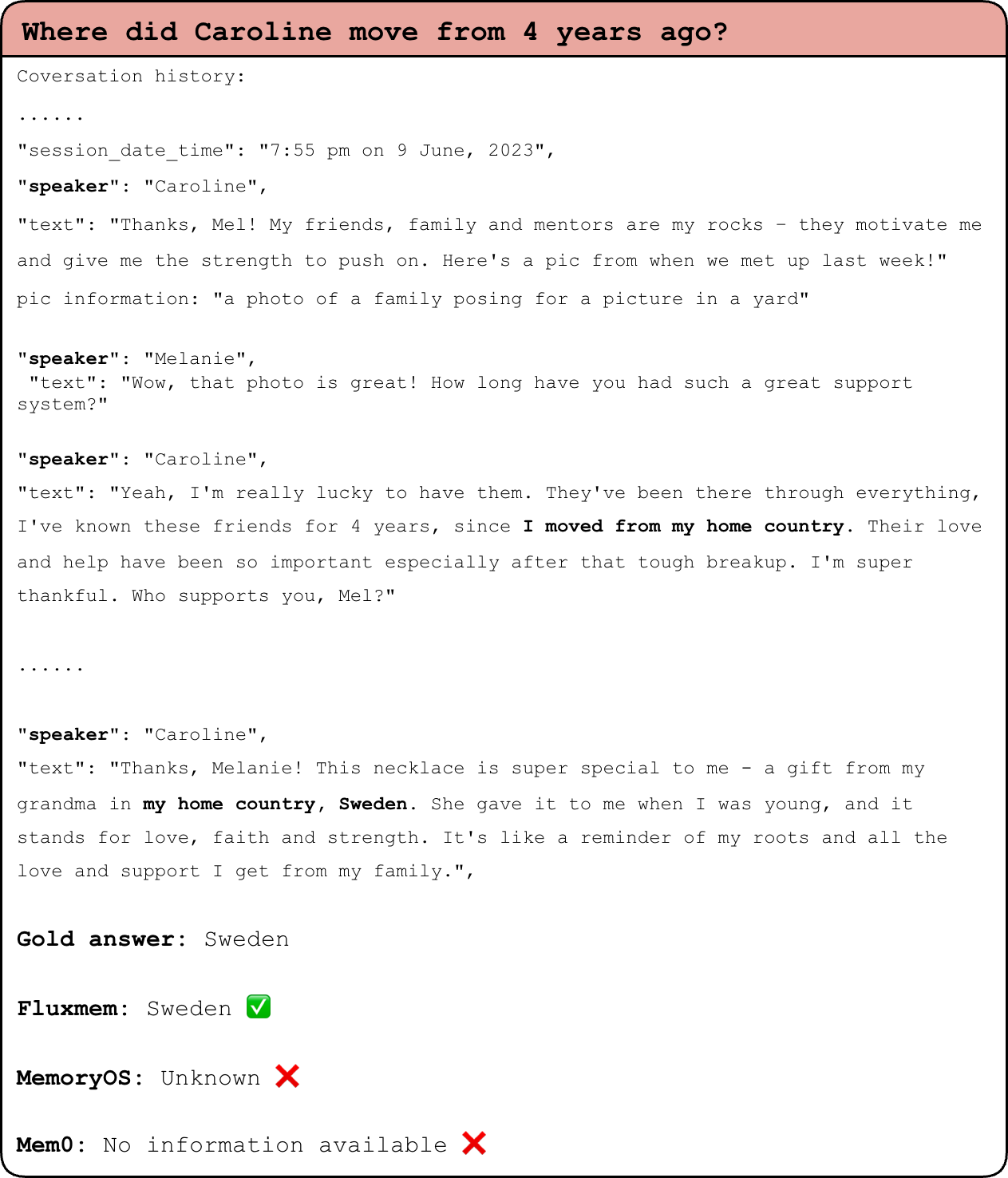}
  \caption{Complex-relation case.}
  \label{figcase2}
\end{figure*}

\begin{figure*}[t]
  \centering
  \includegraphics[width=0.8\linewidth]{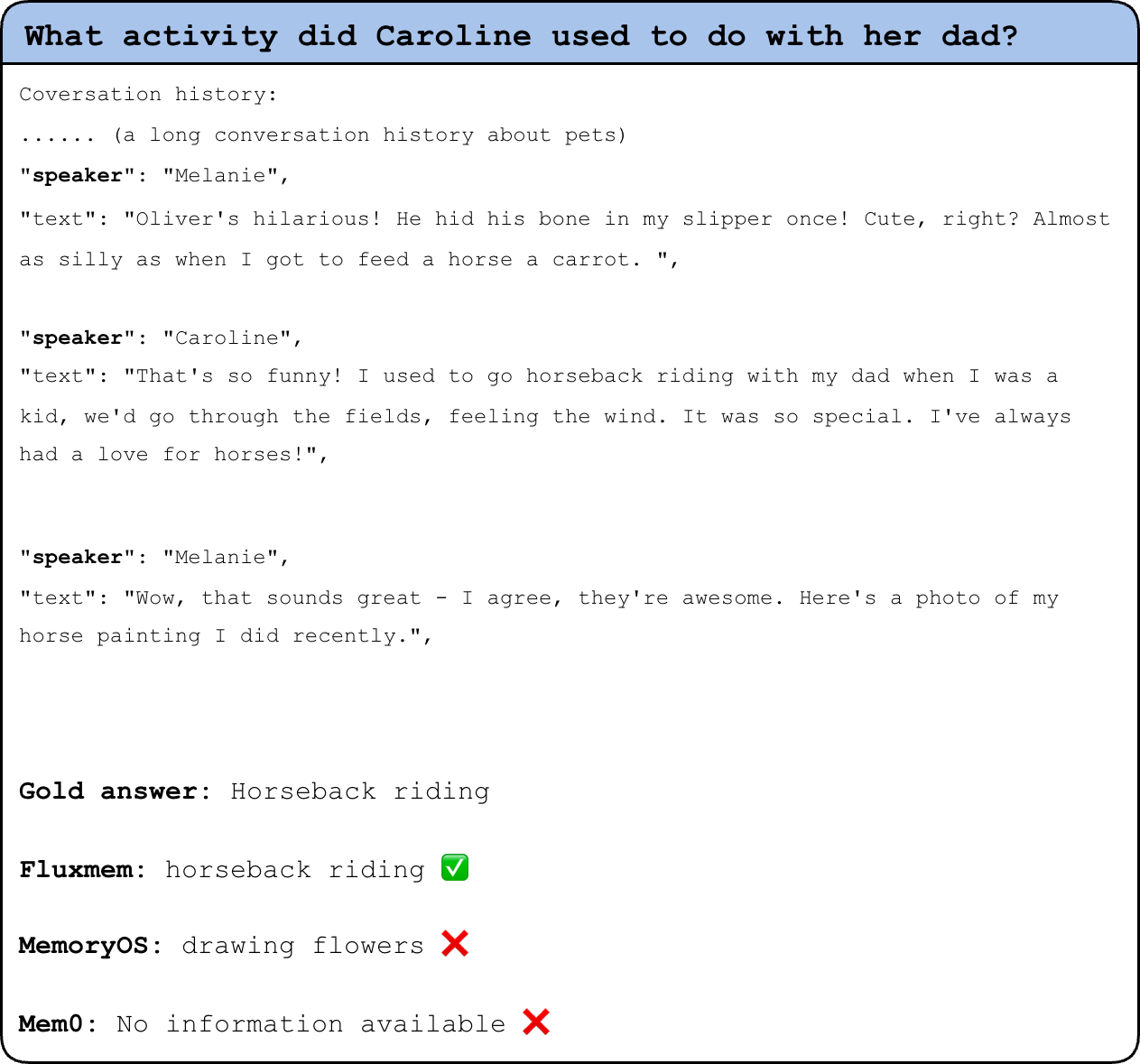}
  \caption{Topic evolution case.}
  \label{figcase3}
\end{figure*}
\FloatBarrier

\section{LLM Prompts Design}
\label{sec:llm_prompts}

\begin{figure*}[h]
  \centering
  \includegraphics[width=0.8\linewidth]{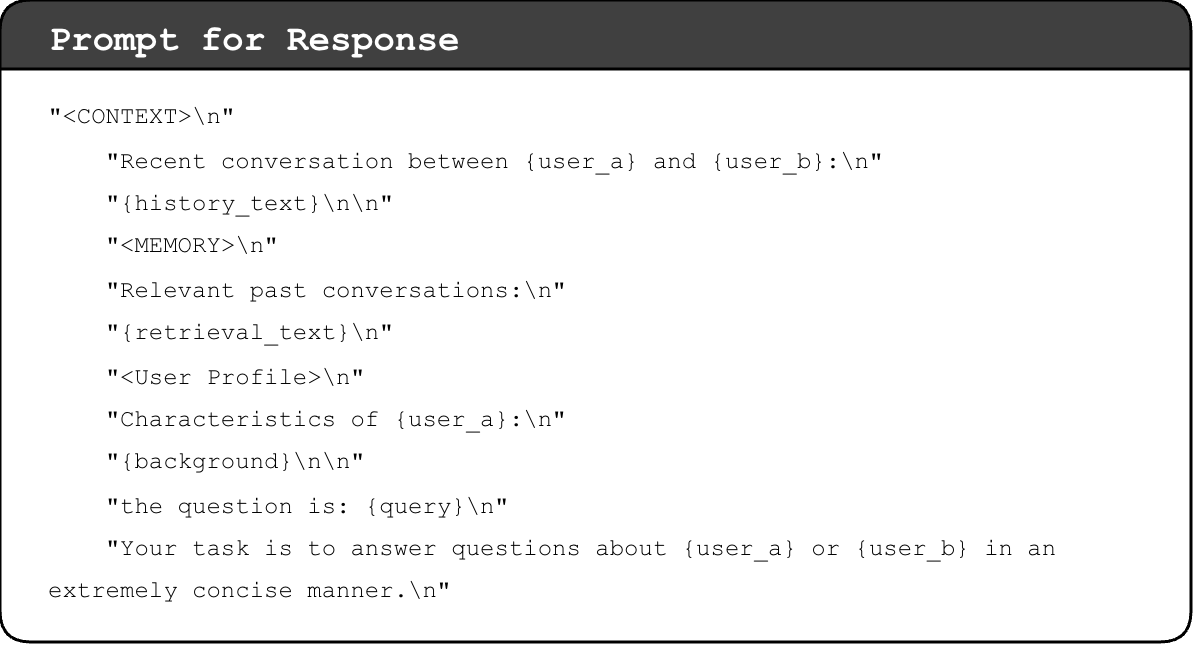}
  \caption{Prompt for response.}
  \label{fig3}
\end{figure*}

\begin{figure*}[h]
  \centering
  \includegraphics[width=0.8\linewidth]{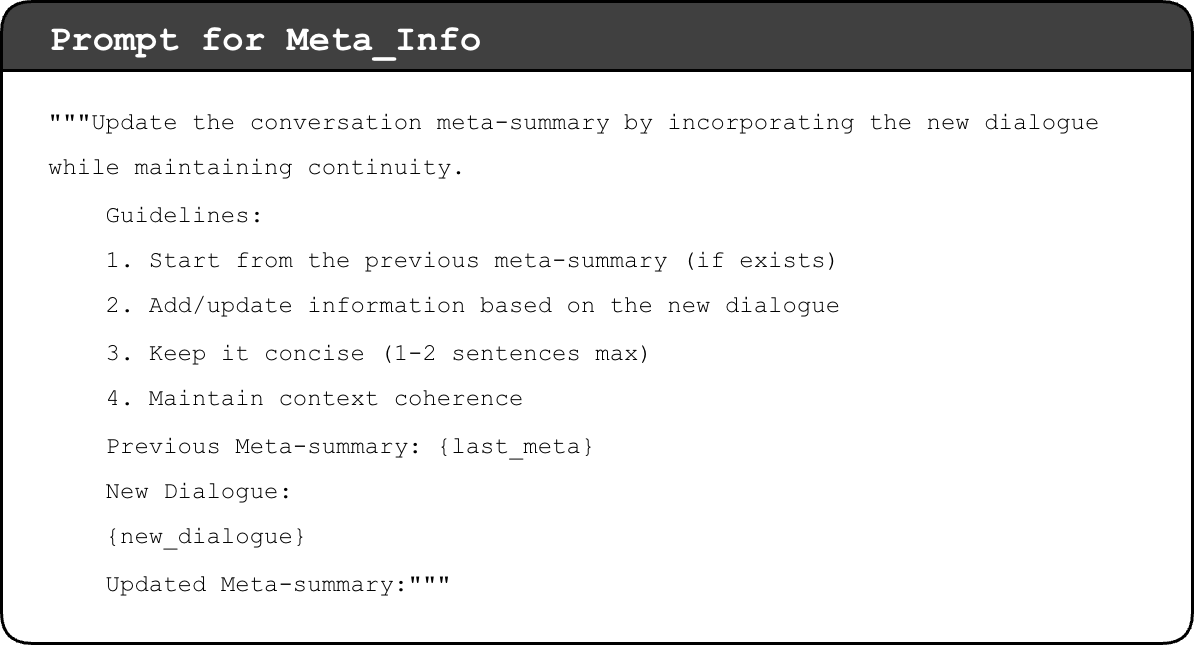}
  \caption{Prompt for Meta\_Info.}
  \label{fig5}
\end{figure*}

\begin{figure*}[t]
  \centering
  \includegraphics[width=0.8\linewidth]{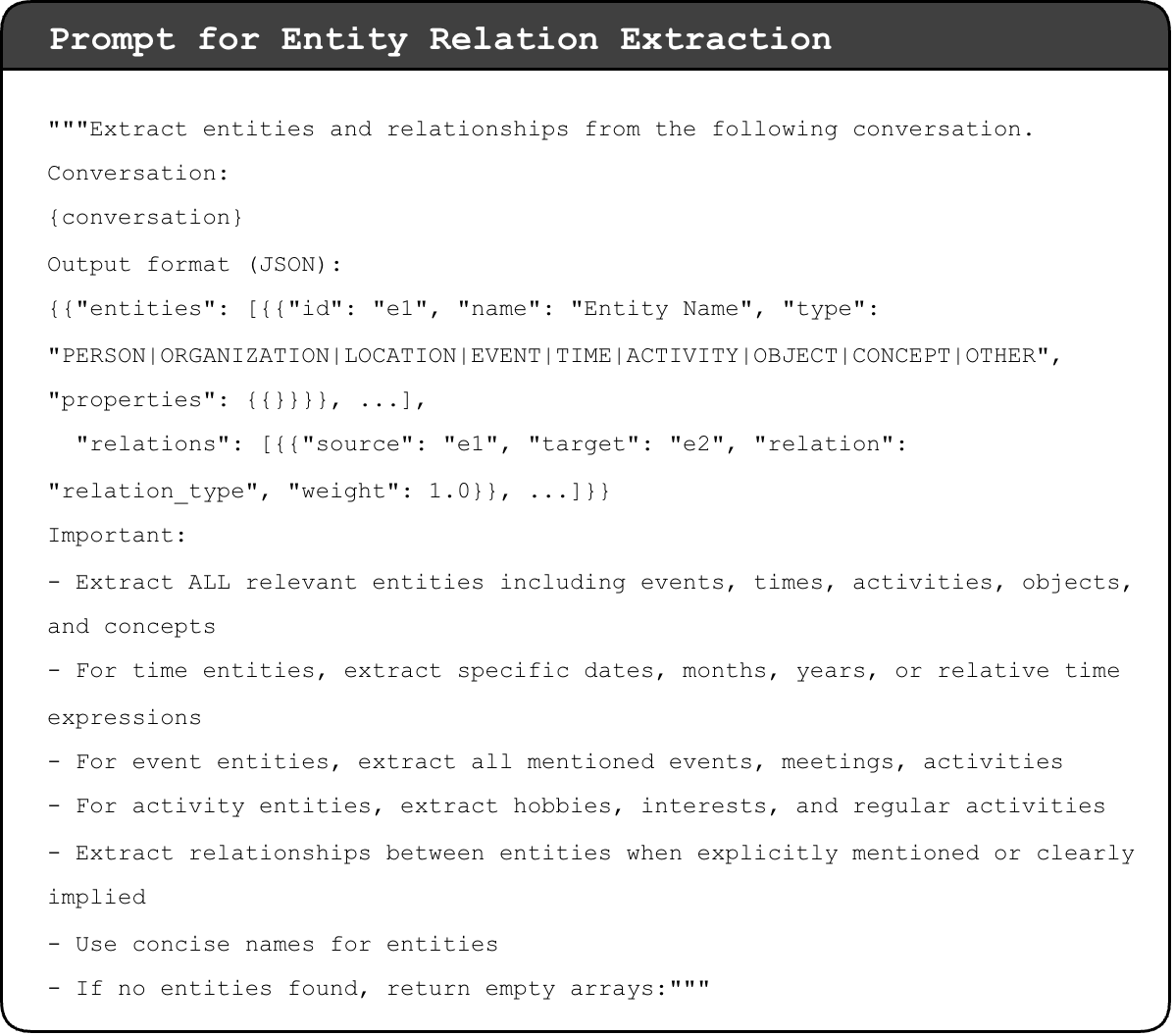}
  \caption{Prompt for entity relation extraction.}
  \label{fig6}
\end{figure*}

\begin{figure*}[h]
  \centering
  \includegraphics[width=0.8\linewidth]{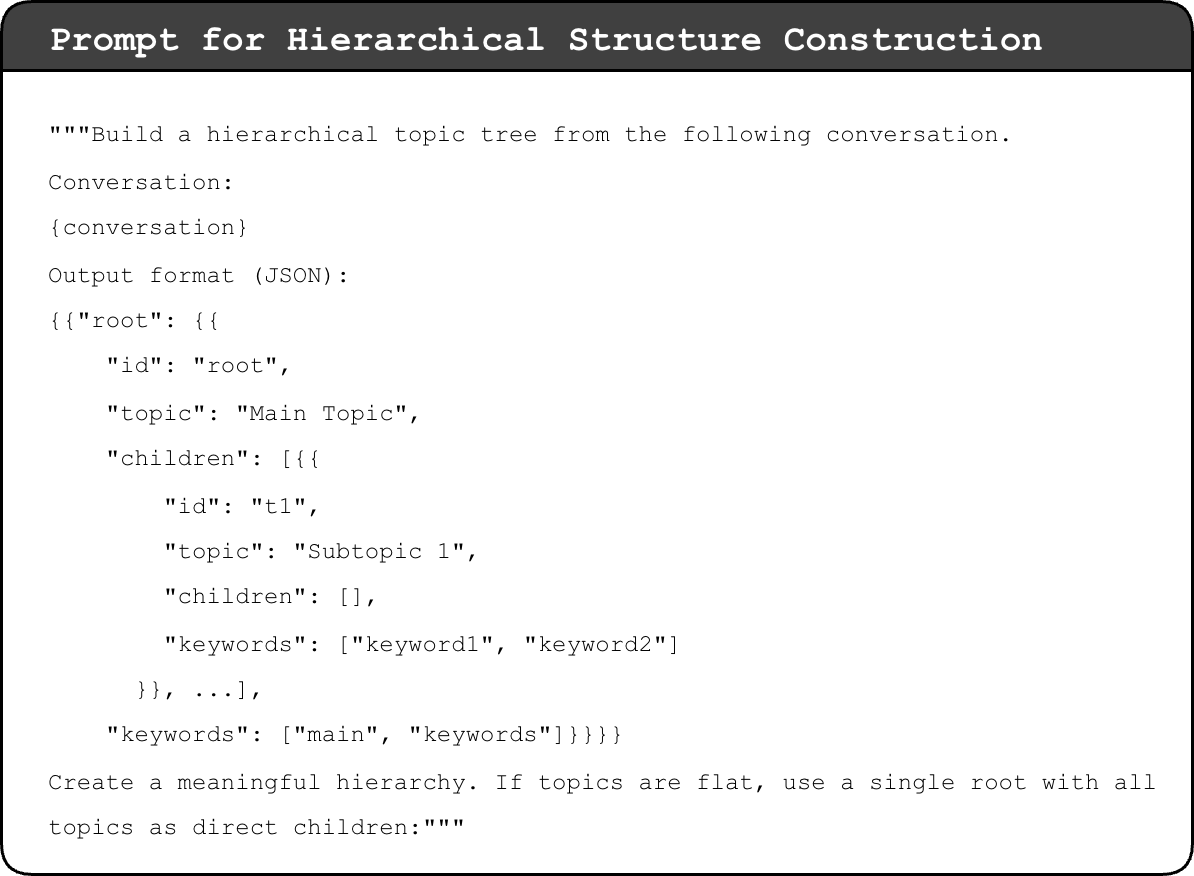}
  \caption{Prompt for hierarchical structure construction.}
  \label{fig7}
\end{figure*}

\begin{figure*}[h]
  \centering
  \includegraphics[width=0.8\linewidth]{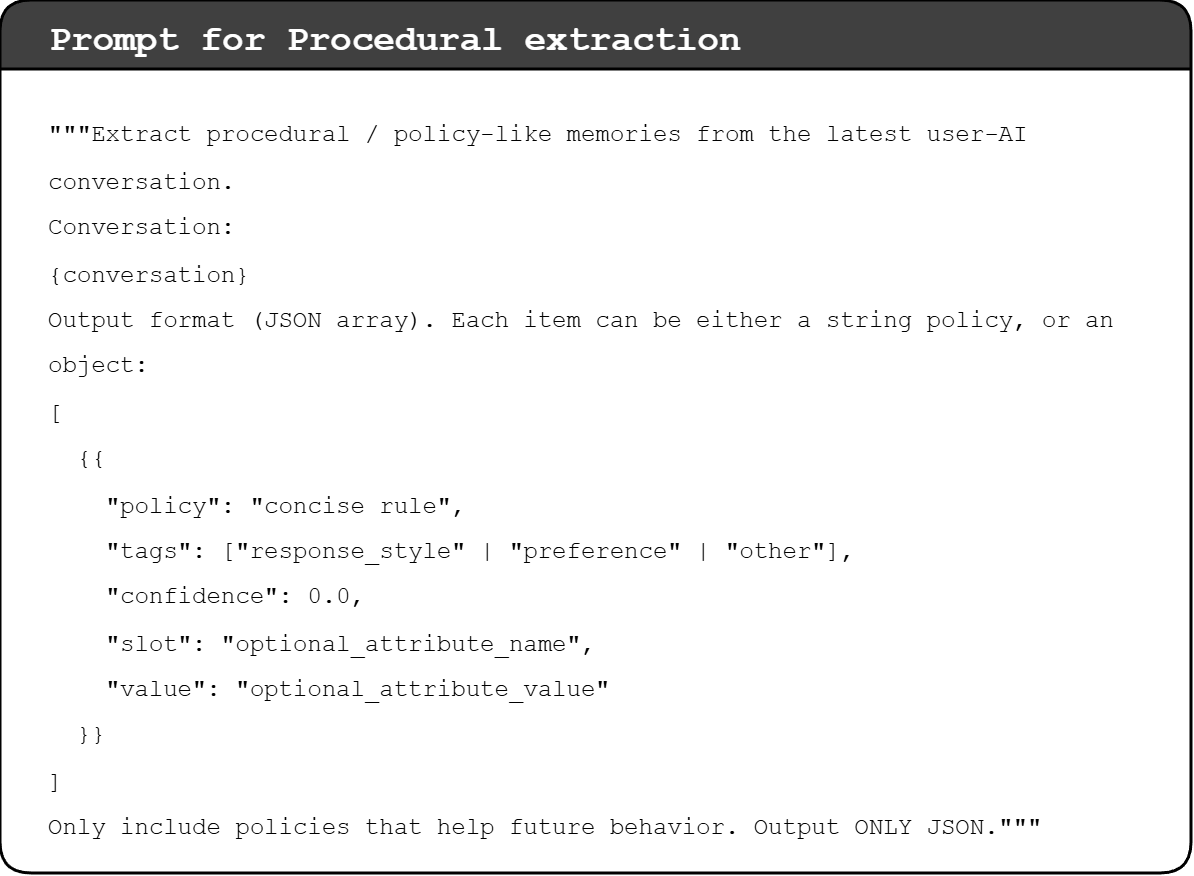}
  \caption{Prompt for procedural extraction.}
  \label{fig8}
\end{figure*}


\end{document}